%% file: main.tex
\documentclass[11pt]{article}
\usepackage[margin=1in]{geometry}
\usepackage{graphicx}
\usepackage{xcolor}
\usepackage{booktabs}
\usepackage{tabularx}
\usepackage{longtable}
\usepackage{placeins}
\newcolumntype{Y}{>{\raggedright\arraybackslash}X}
\usepackage{amsmath}
\usepackage[hidelinks]{hyperref}
\input{macros}
\InputIfFileExists{numbers.tex}{}{}  %

\title{AI Mindmeld: A Latent Interconnect Protocol}
\author{%
  Siavash Golkar\textsuperscript{1,2}\quad
  Jake Kovalic\textsuperscript{1,3}\quad
  Shirley Ho\textsuperscript{1,2,4,5,6}\\[0.75em]
  \small\textsuperscript{1}Polymathic AI\\
  \small\textsuperscript{2}Center for Data Science, New York University\\
  \small\textsuperscript{3}Department of Applied Physics, Yale University\\
  \small\textsuperscript{4}Center for Computational Astrophysics, Flatiron Institute\\
  \small\textsuperscript{5}Department of Astrophysical Sciences, Princeton University\\
  \small\textsuperscript{6}Department of Physics, New York University
}
\date{September 2026}

\begin{document}
\maketitle

\begin{abstract}
\input{sections/abstract}
\end{abstract}

\input{sections/intro}
\input{sections/related}  %
\input{sections/framework}
\input{sections/setup}

\input{sections/results}
\FloatBarrier
\input{sections/discussion}

\bibliographystyle{plain}
\bibliography{refs}

\appendix
\input{sections/appendix}
\end{document}

%% file: macros.tex
\newcommand{\resultdef}[2]{\expandafter\def\csname pla@res@#1\endcsname{#2}}
\newcommand{\result}[1]{\ifcsname pla@res@#1\endcsname\csname pla@res@#1\endcsname\else\PackageError{LIP}{Missing generated result #1}{Regenerate the numerical asset before compiling.}\fi}
\newcommand{\missingbox}[1]{\PackageError{LIP}{Missing generated asset \detokenize{#1}}{Regenerate the asset before compiling.}}
\newcommand{\resultfig}[3][\linewidth]{\begin{figure}[!htbp]\centering\IfFileExists{figures/#2.pdf}{\includegraphics[width=#1]{figures/#2.pdf}}{\missingbox{figures/#2.pdf}}\caption{#3}\label{fig:#2}\end{figure}}
\newcommand{\resulttable}[2]{\begin{table}[t]\centering\caption{#2}\label{tab:#1}\IfFileExists{tables/#1.tex}{\input{tables/#1.tex}}{\missingbox{tables/#1.tex}}\end{table}}

\newcommand{\suppresult}[3]{%
  \IfFileExists{tables/#1.tex}{%
    \begin{table}[!htbp]\centering\small
    \caption{#2}\label{tab:#1}\input{tables/#1.tex}\end{table}%
  }{\missingbox{tables/#1.tex}}%
}

%% file: numbers.tex
\resultdef{compositionMatchedMixedNll}{0.991}
\resultdef{compositionMatchedNllChange}{-0.029}
\resultdef{compositionMatchedNllChangeHi}{-0.014}
\resultdef{compositionMatchedNllChangeLo}{-0.046}
\resultdef{compositionMatchedSameNll}{1.020}
\resultdef{compositionMixedAccPct}{70.3}
\resultdef{compositionOmitFirstPp}{15.9}
\resultdef{compositionOmitSecondPp}{11.9}
\resultdef{compositionPlaceholderAucBenefit}{0.345}
\resultdef{compositionPlaceholderAucBenefitHi}{0.461}
\resultdef{compositionPlaceholderAucBenefitLo}{0.219}
\resultdef{compositionSameAccPct}{85.3}
\resultdef{currentMoleculeAccessMax}{0.93}
\resultdef{currentMoleculeAccessMin}{0.87}
\resultdef{currentMoleculeTransferDrugMax}{0.72}
\resultdef{currentMoleculeTransferDrugMin}{0.65}
\resultdef{currentMoleculeTransferMassMax}{0.93}
\resultdef{currentMoleculeTransferMassMin}{0.89}
\resultdef{currentMoleculeTransferMetaboliteMax}{0.57}
\resultdef{currentMoleculeTransferMetaboliteMin}{0.45}
\resultdef{currentProteinAccessMax}{0.93}
\resultdef{currentProteinAccessMin}{0.92}
\resultdef{currentProteinTransferGroupMax}{0.94}
\resultdef{currentProteinTransferGroupMin}{0.72}
\resultdef{demoNumber}{1.414}
\resultdef{displayLineageAllGroups}{86}
\resultdef{displayLineageAllProteins}{548}
\resultdef{displayLineageRestrictedGroups}{61}
\resultdef{displayLineageRestrictedProteins}{436}
\resultdef{expandedMassMax}{0.972}
\resultdef{expandedMassMin}{0.940}
\resultdef{expandedProteinAccessMax}{0.932}
\resultdef{expandedProteinAccessMin}{0.925}
\resultdef{expandedProteinTransferMax}{0.980}
\resultdef{expandedProteinTransferMin}{0.733}
\resultdef{mergingAdjacentTrained}{0.945}
\resultdef{mergingAdjacentUnseen}{0.895}
\resultdef{mergingBestSecondaryUnseen}{0.859}
\resultdef{mergingBestTrained}{0.932}
\resultdef{mergingBestUnseen}{0.881}
\resultdef{mergingListedSecondaryUnseen}{0.843}
\resultdef{mergingListedTrained}{0.953}
\resultdef{mergingListedUnseen}{0.894}
\resultdef{methodMolCropped}{91}
\resultdef{methodMolEmbedded}{45,929}
\resultdef{methodMolFailed}{41}
\resultdef{methodMolNoSmiles}{43}
\resultdef{methodMolTotal}{46,013}
\resultdef{molAccessCaptionGapChemfm}{0.340}
\resultdef{molAccessCaptionGapMolformer}{0.282}
\resultdef{molAccessCaptionGapUnimol}{0.315}
\resultdef{molAccessFullDeltaChemfm}{+0.106}
\resultdef{molAccessFullDeltaHiChemfm}{+0.141}
\resultdef{molAccessFullDeltaHiMolformer}{+0.174}
\resultdef{molAccessFullDeltaHiUnimol}{+0.145}
\resultdef{molAccessFullDeltaLoChemfm}{+0.073}
\resultdef{molAccessFullDeltaLoMolformer}{+0.111}
\resultdef{molAccessFullDeltaLoUnimol}{+0.088}
\resultdef{molAccessFullDeltaMolformer}{+0.138}
\resultdef{molAccessFullDeltaUnimol}{+0.112}
\resultdef{molAccessFullOwnChemfm}{0.899}
\resultdef{molAccessFullOwnMolformer}{0.902}
\resultdef{molAccessFullOwnUnimol}{0.900}
\resultdef{molAccessFullShufChemfm}{0.497}
\resultdef{molAccessFullShufMolformer}{0.543}
\resultdef{molAccessFullShufUnimol}{0.513}
\resultdef{molAccessGapEightChemfm}{1.14}
\resultdef{molAccessGapEightMolformer}{0.91}
\resultdef{molAccessGapEightUnimol}{1.01}
\resultdef{molAccessScanDeltaChemfm}{+0.127}
\resultdef{molAccessScanDeltaHiChemfm}{+0.166}
\resultdef{molAccessScanDeltaHiMolformer}{+0.213}
\resultdef{molAccessScanDeltaHiUnimol}{+0.193}
\resultdef{molAccessScanDeltaLoChemfm}{+0.088}
\resultdef{molAccessScanDeltaLoMolformer}{+0.111}
\resultdef{molAccessScanDeltaLoUnimol}{+0.107}
\resultdef{molAccessScanDeltaMolformer}{+0.162}
\resultdef{molAccessScanDeltaUnimol}{+0.153}
\resultdef{molAccessScanOwnChemfm}{0.968}
\resultdef{molAccessScanOwnMolformer}{0.933}
\resultdef{molAccessScanOwnUnimol}{0.940}
\resultdef{molAccessScanShufChemfm}{0.526}
\resultdef{molAccessScanShufMolformer}{0.529}
\resultdef{molAccessScanShufUnimol}{0.505}
\resultdef{molChemfmMetaboliteFinal}{0.629}
\resultdef{molChemfmMetaboliteInit}{0.764}
\resultdef{molPopEvalN}{200}
\resultdef{molPopFullClasses}{24}
\resultdef{molPopMassAllPos}{89}
\resultdef{molPopMassUnstatedN}{177}
\resultdef{molPopMassUnstatedPos}{67}
\resultdef{molPopRelHeldoutItems}{48}
\resultdef{molPopRelItems}{79}
\resultdef{molPopRelMixedItems}{31}
\resultdef{molPopRoleTasks}{29}
\resultdef{molPopRoleTasksLowPos}{19}
\resultdef{molPopScanClasses}{5}
\resultdef{molPopUnionN}{821}
\resultdef{molTransferMassAllDeltaChemfm}{+0.412}
\resultdef{molTransferMassAllDeltaHiChemfm}{+0.493}
\resultdef{molTransferMassAllDeltaHiMolformer}{+0.523}
\resultdef{molTransferMassAllDeltaHiUnimol}{+0.520}
\resultdef{molTransferMassAllDeltaLoChemfm}{+0.331}
\resultdef{molTransferMassAllDeltaLoMolformer}{+0.374}
\resultdef{molTransferMassAllDeltaLoUnimol}{+0.358}
\resultdef{molTransferMassAllDeltaMolformer}{+0.446}
\resultdef{molTransferMassAllDeltaUnimol}{+0.439}
\resultdef{molTransferMassAllOwnChemfm}{0.964}
\resultdef{molTransferMassAllOwnMolformer}{0.975}
\resultdef{molTransferMassAllOwnUnimol}{0.993}
\resultdef{molTransferMassUnstatedDeltaChemfm}{+0.378}
\resultdef{molTransferMassUnstatedDeltaHiChemfm}{+0.463}
\resultdef{molTransferMassUnstatedDeltaHiMolformer}{+0.475}
\resultdef{molTransferMassUnstatedDeltaHiUnimol}{+0.497}
\resultdef{molTransferMassUnstatedDeltaLoChemfm}{+0.295}
\resultdef{molTransferMassUnstatedDeltaLoMolformer}{+0.303}
\resultdef{molTransferMassUnstatedDeltaLoUnimol}{+0.323}
\resultdef{molTransferMassUnstatedDeltaMolformer}{+0.388}
\resultdef{molTransferMassUnstatedDeltaUnimol}{+0.411}
\resultdef{molTransferMassUnstatedOwnChemfm}{0.953}
\resultdef{molTransferMassUnstatedOwnMolformer}{0.970}
\resultdef{molTransferMassUnstatedOwnUnimol}{0.990}
\resultdef{molTransferRoleAllDeltaChemfm}{-0.056}
\resultdef{molTransferRoleAllDeltaHiChemfm}{+0.023}
\resultdef{molTransferRoleAllDeltaHiMolformer}{+0.084}
\resultdef{molTransferRoleAllDeltaHiUnimol}{+0.162}
\resultdef{molTransferRoleAllDeltaLoChemfm}{-0.133}
\resultdef{molTransferRoleAllDeltaLoMolformer}{-0.109}
\resultdef{molTransferRoleAllDeltaLoUnimol}{-0.010}
\resultdef{molTransferRoleAllDeltaMolformer}{-0.000}
\resultdef{molTransferRoleAllDeltaUnimol}{+0.079}
\resultdef{molTransferRoleAllOwnChemfm}{0.691}
\resultdef{molTransferRoleAllOwnMolformer}{0.666}
\resultdef{molTransferRoleAllOwnUnimol}{0.712}
\resultdef{molTransferRoleUnstatedDeltaChemfm}{-0.047}
\resultdef{molTransferRoleUnstatedDeltaHiChemfm}{+0.046}
\resultdef{molTransferRoleUnstatedDeltaHiMolformer}{+0.069}
\resultdef{molTransferRoleUnstatedDeltaHiUnimol}{+0.194}
\resultdef{molTransferRoleUnstatedDeltaLoChemfm}{-0.136}
\resultdef{molTransferRoleUnstatedDeltaLoMolformer}{-0.133}
\resultdef{molTransferRoleUnstatedDeltaLoUnimol}{+0.012}
\resultdef{molTransferRoleUnstatedDeltaMolformer}{-0.019}
\resultdef{molTransferRoleUnstatedDeltaUnimol}{+0.101}
\resultdef{molTransferRoleUnstatedOwnChemfm}{0.689}
\resultdef{molTransferRoleUnstatedOwnMolformer}{0.667}
\resultdef{molTransferRoleUnstatedOwnUnimol}{0.716}
\resultdef{molTransferUnionRoleDeltaChemfm}{-0.016}
\resultdef{molTransferUnionRoleDeltaHiChemfm}{+0.018}
\resultdef{molTransferUnionRoleDeltaHiMolformer}{+0.035}
\resultdef{molTransferUnionRoleDeltaHiUnimol}{+0.067}
\resultdef{molTransferUnionRoleDeltaLoChemfm}{-0.052}
\resultdef{molTransferUnionRoleDeltaLoMolformer}{-0.055}
\resultdef{molTransferUnionRoleDeltaLoUnimol}{-0.010}
\resultdef{molTransferUnionRoleDeltaMolformer}{-0.009}
\resultdef{molTransferUnionRoleDeltaUnimol}{+0.028}
\resultdef{molTransferUnionRoleNameGapChemfm}{-0.138}
\resultdef{molTransferUnionRoleNameGapHiChemfm}{-0.106}
\resultdef{molTransferUnionRoleNameGapHiMolformer}{-0.174}
\resultdef{molTransferUnionRoleNameGapHiUnimol}{-0.151}
\resultdef{molTransferUnionRoleNameGapLoChemfm}{-0.171}
\resultdef{molTransferUnionRoleNameGapLoMolformer}{-0.265}
\resultdef{molTransferUnionRoleNameGapLoUnimol}{-0.228}
\resultdef{molTransferUnionRoleNameGapMolformer}{-0.219}
\resultdef{molTransferUnionRoleNameGapUnimol}{-0.190}
\resultdef{molTransferUnionRoleOwnChemfm}{0.726}
\resultdef{molTransferUnionRoleOwnMolformer}{0.645}
\resultdef{molTransferUnionRoleOwnUnimol}{0.673}
\resultdef{moleculeTransferRoleGroupMax}{0.817}
\resultdef{moleculeTransferRoleGroupMin}{0.555}
\resultdef{nameRecoveryOverallPercent}{13.2}
\resultdef{pAccCapGapNameEsmc}{0.322}
\resultdef{pAccCapGapNameHiEsmc}{0.393}
\resultdef{pAccCapGapNameHiMimic}{0.492}
\resultdef{pAccCapGapNameHiPglm}{0.374}
\resultdef{pAccCapGapNameLoEsmc}{0.254}
\resultdef{pAccCapGapNameLoMimic}{0.340}
\resultdef{pAccCapGapNameLoPglm}{0.226}
\resultdef{pAccCapGapNameMimic}{0.411}
\resultdef{pAccCapGapNamePglm}{0.297}
\resultdef{pAccCapGapPlhEsmc}{0.621}
\resultdef{pAccCapGapPlhHiEsmc}{0.723}
\resultdef{pAccCapGapPlhHiMimic}{0.826}
\resultdef{pAccCapGapPlhHiPglm}{0.707}
\resultdef{pAccCapGapPlhLoEsmc}{0.524}
\resultdef{pAccCapGapPlhLoMimic}{0.610}
\resultdef{pAccCapGapPlhLoPglm}{0.503}
\resultdef{pAccCapGapPlhMimic}{0.716}
\resultdef{pAccCapGapPlhPglm}{0.603}
\resultdef{pAccCapGapShufEsmc}{0.310}
\resultdef{pAccCapGapShufHiEsmc}{0.395}
\resultdef{pAccCapGapShufHiMimic}{0.372}
\resultdef{pAccCapGapShufHiPglm}{0.427}
\resultdef{pAccCapGapShufLoEsmc}{0.231}
\resultdef{pAccCapGapShufLoMimic}{0.246}
\resultdef{pAccCapGapShufLoPglm}{0.250}
\resultdef{pAccCapGapShufMimic}{0.305}
\resultdef{pAccCapGapShufPglm}{0.336}
\resultdef{pAccKwOwnEsmc}{0.936}
\resultdef{pAccKwOwnMimic}{0.935}
\resultdef{pAccKwOwnMinusShufEsmc}{+0.385}
\resultdef{pAccKwOwnMinusShufHiEsmc}{+0.447}
\resultdef{pAccKwOwnMinusShufHiMimic}{+0.489}
\resultdef{pAccKwOwnMinusShufHiPglm}{+0.487}
\resultdef{pAccKwOwnMinusShufLoEsmc}{+0.321}
\resultdef{pAccKwOwnMinusShufLoMimic}{+0.332}
\resultdef{pAccKwOwnMinusShufLoPglm}{+0.335}
\resultdef{pAccKwOwnMinusShufMimic}{+0.414}
\resultdef{pAccKwOwnMinusShufPglm}{+0.412}
\resultdef{pAccKwOwnPglm}{0.945}
\resultdef{pAccKwShufEsmc}{0.551}
\resultdef{pAccKwShufMimic}{0.520}
\resultdef{pAccKwShufPglm}{0.533}
\resultdef{pAccKwTasksEval}{5}
\resultdef{pAccKwTasksUnion}{14}
\resultdef{pAccNEval}{200}
\resultdef{pAccNUnion}{2,036}
\resultdef{pAccUnionKwNameEsmc}{0.943}
\resultdef{pAccUnionKwNameMimic}{0.943}
\resultdef{pAccUnionKwNamePglm}{0.943}
\resultdef{pAccUnionKwOwnEsmc}{0.911}
\resultdef{pAccUnionKwOwnMimic}{0.902}
\resultdef{pAccUnionKwOwnMinusNameEsmc}{-0.032}
\resultdef{pAccUnionKwOwnMinusNameHiEsmc}{-0.026}
\resultdef{pAccUnionKwOwnMinusNameHiMimic}{-0.034}
\resultdef{pAccUnionKwOwnMinusNameHiPglm}{-0.018}
\resultdef{pAccUnionKwOwnMinusNameLoEsmc}{-0.038}
\resultdef{pAccUnionKwOwnMinusNameLoMimic}{-0.048}
\resultdef{pAccUnionKwOwnMinusNameLoPglm}{-0.030}
\resultdef{pAccUnionKwOwnMinusNameMimic}{-0.041}
\resultdef{pAccUnionKwOwnMinusNamePglm}{-0.024}
\resultdef{pAccUnionKwOwnPglm}{0.919}
\resultdef{pCovHeldoutOwnEsmc}{0.862}
\resultdef{pCovHeldoutOwnMimic}{0.860}
\resultdef{pCovHeldoutOwnMinusNameEsmc}{-0.004}
\resultdef{pCovHeldoutOwnMinusNameHiEsmc}{+0.006}
\resultdef{pCovHeldoutOwnMinusNameHiMimic}{+0.024}
\resultdef{pCovHeldoutOwnMinusNameHiPglm}{+0.022}
\resultdef{pCovHeldoutOwnMinusNameLoEsmc}{-0.014}
\resultdef{pCovHeldoutOwnMinusNameLoMimic}{-0.019}
\resultdef{pCovHeldoutOwnMinusNameLoPglm}{-0.021}
\resultdef{pCovHeldoutOwnMinusNameMimic}{+0.002}
\resultdef{pCovHeldoutOwnMinusNamePglm}{+0.000}
\resultdef{pCovHeldoutOwnMinusPlhEsmc}{+0.360}
\resultdef{pCovHeldoutOwnMinusPlhMimic}{+0.371}
\resultdef{pCovHeldoutOwnMinusPlhPglm}{+0.369}
\resultdef{pCovHeldoutOwnPglm}{0.859}
\resultdef{pCovHeldoutTasksEsmc}{17}
\resultdef{pCovHeldoutTasksMimic}{10}
\resultdef{pCovHeldoutTasksPglm}{10}
\resultdef{pCovHomGradientEsmc}{+0.010}
\resultdef{pCovHomGradientHiEsmc}{+0.033}
\resultdef{pCovHomGradientHiMimic}{+0.039}
\resultdef{pCovHomGradientHiPglm}{+0.017}
\resultdef{pCovHomGradientLoEsmc}{-0.014}
\resultdef{pCovHomGradientLoMimic}{-0.002}
\resultdef{pCovHomGradientLoPglm}{-0.022}
\resultdef{pCovHomGradientMimic}{+0.018}
\resultdef{pCovHomGradientPglm}{-0.002}
\resultdef{pCovHomGradientReader}{+0.036}
\resultdef{pCovHomNoHitN}{747}
\resultdef{pCovLenMcAccName}{0.526}
\resultdef{pCovLenMcAccOwn}{0.531}
\resultdef{pCovLenMcSpearmanName}{0.776}
\resultdef{pCovLenMcSpearmanOwn}{0.845}
\resultdef{pCovReaderBeatsOwnTasksEsmc}{10}
\resultdef{pCovReaderBeatsOwnTasksMimic}{13}
\resultdef{pCovReaderBeatsOwnTasksPglm}{11}
\resultdef{pCovSeqOwnEsmc}{0.906}
\resultdef{pCovSeqOwnMimic}{0.940}
\resultdef{pCovSeqOwnMinusNameEsmc}{+0.054}
\resultdef{pCovSeqOwnMinusNameHiEsmc}{+0.072}
\resultdef{pCovSeqOwnMinusNameHiMimic}{+0.086}
\resultdef{pCovSeqOwnMinusNameHiPglm}{+0.026}
\resultdef{pCovSeqOwnMinusNameLoEsmc}{+0.036}
\resultdef{pCovSeqOwnMinusNameLoMimic}{+0.030}
\resultdef{pCovSeqOwnMinusNameLoPglm}{-0.057}
\resultdef{pCovSeqOwnMinusNameMimic}{+0.058}
\resultdef{pCovSeqOwnMinusNamePglm}{-0.013}
\resultdef{pCovSeqOwnMinusPlhEsmc}{+0.394}
\resultdef{pCovSeqOwnMinusPlhMimic}{+0.405}
\resultdef{pCovSeqOwnMinusPlhPglm}{+0.334}
\resultdef{pCovSeqOwnPglm}{0.868}
\resultdef{pCovSeqTasksEsmc}{3}
\resultdef{pCovSeqTasksMimic}{1}
\resultdef{pCovSeqTasksPglm}{1}
\resultdef{pCovTrainedFieldOwnEsmc}{0.945}
\resultdef{pCovTrainedFieldOwnMimic}{0.958}
\resultdef{pCovTrainedFieldOwnMinusNameEsmc}{-0.008}
\resultdef{pCovTrainedFieldOwnMinusNameHiEsmc}{+0.001}
\resultdef{pCovTrainedFieldOwnMinusNameHiMimic}{-0.004}
\resultdef{pCovTrainedFieldOwnMinusNameHiPglm}{-0.005}
\resultdef{pCovTrainedFieldOwnMinusNameLoEsmc}{-0.016}
\resultdef{pCovTrainedFieldOwnMinusNameLoMimic}{-0.026}
\resultdef{pCovTrainedFieldOwnMinusNameLoPglm}{-0.032}
\resultdef{pCovTrainedFieldOwnMinusNameMimic}{-0.015}
\resultdef{pCovTrainedFieldOwnMinusNamePglm}{-0.018}
\resultdef{pCovTrainedFieldOwnMinusPlhEsmc}{+0.442}
\resultdef{pCovTrainedFieldOwnMinusPlhMimic}{+0.471}
\resultdef{pCovTrainedFieldOwnMinusPlhPglm}{+0.468}
\resultdef{pCovTrainedFieldOwnPglm}{0.956}
\resultdef{pCovTrainedFieldTasksEsmc}{6}
\resultdef{pCovTrainedFieldTasksMimic}{2}
\resultdef{pCovTrainedFieldTasksPglm}{2}
\resultdef{pExactLengthGroupsMimic}{5}
\resultdef{pExactLengthPairsMimic}{27}
\resultdef{pExactLineageConcEsmc}{0.789}
\resultdef{pExactLineageOwnMinusHalfEsmc}{+0.289}
\resultdef{pExactLineageOwnMinusHalfHiEsmc}{+0.319}
\resultdef{pExactLineageOwnMinusHalfLoEsmc}{+0.246}
\resultdef{pExactLineageOwnMinusNameEsmc}{+0.314}
\resultdef{pExactLineageOwnMinusNameHiEsmc}{+0.376}
\resultdef{pExactLineageOwnMinusNameLoEsmc}{+0.269}
\resultdef{pExactNameGroups}{201}
\resultdef{pExactNameProteins}{858}
\resultdef{pExactNameVerified}{2,036}
\resultdef{pExactTThreeLineageConc}{0.789}
\resultdef{pExtDeepLocBinaryEsmcStepFiveHundred}{0.920}
\resultdef{pExtDeepLocBinaryEsmcStepThousand}{0.906}
\resultdef{pExtDeepLocBinaryMimicStepFiveHundred}{0.912}
\resultdef{pExtDeepLocBinaryMimicStepThousand}{0.910}
\resultdef{pExtDeepLocBinaryPglmStepFiveHundred}{0.913}
\resultdef{pExtDeepLocTwoEsmcStepFiveHundred}{0.852}
\resultdef{pExtDeepLocTwoEsmcStepThousand}{0.845}
\resultdef{pExtDeepLocTwoHiEsmcStepFiveHundred}{0.873}
\resultdef{pExtDeepLocTwoHiEsmcStepThousand}{0.869}
\resultdef{pExtDeepLocTwoHiMimicStepFiveHundred}{0.849}
\resultdef{pExtDeepLocTwoHiMimicStepThousand}{0.848}
\resultdef{pExtDeepLocTwoHiPglmStepFiveHundred}{0.863}
\resultdef{pExtDeepLocTwoLoEsmcStepFiveHundred}{0.829}
\resultdef{pExtDeepLocTwoLoEsmcStepThousand}{0.816}
\resultdef{pExtDeepLocTwoLoMimicStepFiveHundred}{0.799}
\resultdef{pExtDeepLocTwoLoMimicStepThousand}{0.803}
\resultdef{pExtDeepLocTwoLoPglmStepFiveHundred}{0.821}
\resultdef{pExtDeepLocTwoMimicStepFiveHundred}{0.826}
\resultdef{pExtDeepLocTwoMimicStepThousand}{0.826}
\resultdef{pExtDeepLocTwoPglmStepFiveHundred}{0.843}
\resultdef{pExtEngMeanEsmcStepFiveHundred}{0.537}
\resultdef{pExtEngMeanEsmcStepThousand}{0.510}
\resultdef{pExtEngMeanHiEsmcStepFiveHundred}{0.557}
\resultdef{pExtEngMeanHiEsmcStepThousand}{0.529}
\resultdef{pExtEngMeanHiMimicStepFiveHundred}{0.538}
\resultdef{pExtEngMeanHiMimicStepThousand}{0.536}
\resultdef{pExtEngMeanHiPglmStepFiveHundred}{0.553}
\resultdef{pExtEngMeanLoEsmcStepFiveHundred}{0.517}
\resultdef{pExtEngMeanLoEsmcStepThousand}{0.490}
\resultdef{pExtEngMeanLoMimicStepFiveHundred}{0.498}
\resultdef{pExtEngMeanLoMimicStepThousand}{0.498}
\resultdef{pExtEngMeanLoPglmStepFiveHundred}{0.513}
\resultdef{pExtEngMeanMimicStepFiveHundred}{0.518}
\resultdef{pExtEngMeanMimicStepThousand}{0.517}
\resultdef{pExtEngMeanPglmStepFiveHundred}{0.534}
\resultdef{pGuessGuessEsmc}{0.819}
\resultdef{pGuessGuessMimic}{0.790}
\resultdef{pGuessGuessPglm}{0.795}
\resultdef{pGuessHeldoutOwnMinusGuessEsmc}{+0.061}
\resultdef{pGuessHeldoutOwnMinusGuessHiEsmc}{+0.070}
\resultdef{pGuessHeldoutOwnMinusGuessHiMimic}{+0.114}
\resultdef{pGuessHeldoutOwnMinusGuessHiPglm}{+0.100}
\resultdef{pGuessHeldoutOwnMinusGuessLoEsmc}{+0.052}
\resultdef{pGuessHeldoutOwnMinusGuessLoMimic}{+0.066}
\resultdef{pGuessHeldoutOwnMinusGuessLoPglm}{+0.057}
\resultdef{pGuessHeldoutOwnMinusGuessMimic}{+0.089}
\resultdef{pGuessHeldoutOwnMinusGuessPglm}{+0.078}
\resultdef{pGuessMinusPlhEsmc}{+0.315}
\resultdef{pGuessMinusPlhMimic}{+0.297}
\resultdef{pGuessMinusPlhPglm}{+0.303}
\resultdef{pGuessOwnEsmc}{0.887}
\resultdef{pGuessOwnMimic}{0.881}
\resultdef{pGuessOwnMinusGuessEsmc}{+0.068}
\resultdef{pGuessOwnMinusGuessHiEsmc}{+0.076}
\resultdef{pGuessOwnMinusGuessHiMimic}{+0.112}
\resultdef{pGuessOwnMinusGuessHiPglm}{+0.096}
\resultdef{pGuessOwnMinusGuessLoEsmc}{+0.060}
\resultdef{pGuessOwnMinusGuessLoMimic}{+0.072}
\resultdef{pGuessOwnMinusGuessLoPglm}{+0.062}
\resultdef{pGuessOwnMinusGuessMimic}{+0.091}
\resultdef{pGuessOwnMinusGuessPglm}{+0.079}
\resultdef{pGuessOwnPglm}{0.874}
\resultdef{pGuessProteinsEsmc}{2,036}
\resultdef{pGuessProteinsMimic}{600}
\resultdef{pGuessProteinsPglm}{600}
\resultdef{pGuessTasksEsmc}{26}
\resultdef{pGuessTasksMimic}{13}
\resultdef{pGuessTasksPglm}{13}
\resultdef{pInjobGramProteins}{100}
\resultdef{pInjobGramSeenGapEsmc}{0.352}
\resultdef{pInjobGramSeenGapMimic}{0.350}
\resultdef{pInjobGramSeenGapPglm}{0.340}
\resultdef{pInjobGramUnseenGapEsmc}{0.358}
\resultdef{pInjobGramUnseenGapMimic}{0.360}
\resultdef{pInjobGramUnseenGapPglm}{0.361}
\resultdef{pInjobInitGramSeenGapEsmc}{0.082}
\resultdef{pInjobInitGramSeenGapMimic}{0.064}
\resultdef{pInjobInitGramSeenGapPglm}{0.071}
\resultdef{pInjobInitGramUnseenGapEsmc}{0.071}
\resultdef{pInjobInitGramUnseenGapMimic}{0.064}
\resultdef{pInjobInitGramUnseenGapPglm}{0.083}
\resultdef{pInjobInitMultiHeldoutGapEsmc}{0.069}
\resultdef{pInjobInitMultiHeldoutGapMimic}{0.066}
\resultdef{pInjobInitMultiHeldoutGapPglm}{0.079}
\resultdef{pInjobInitMultiTrainGapEsmc}{0.091}
\resultdef{pInjobInitMultiTrainGapMimic}{0.093}
\resultdef{pInjobInitMultiTrainGapPglm}{0.102}
\resultdef{pInjobInitRelHeldoutGapEsmc}{0.033}
\resultdef{pInjobInitRelHeldoutGapMimic}{0.068}
\resultdef{pInjobInitRelHeldoutGapPglm}{0.074}
\resultdef{pInjobInitRelMixedGapEsmc}{0.064}
\resultdef{pInjobInitRelMixedGapMimic}{0.098}
\resultdef{pInjobInitRelMixedGapPglm}{0.112}
\resultdef{pInjobInitRelvHeldoutGapEsmc}{0.042}
\resultdef{pInjobInitRelvHeldoutGapMimic}{0.088}
\resultdef{pInjobInitRelvHeldoutGapPglm}{0.096}
\resultdef{pInjobInitRelvMixedGapEsmc}{0.080}
\resultdef{pInjobInitRelvMixedGapMimic}{0.116}
\resultdef{pInjobInitRelvMixedGapPglm}{0.128}
\resultdef{pInjobMultiHeldoutGapEsmc}{0.317}
\resultdef{pInjobMultiHeldoutGapMimic}{0.355}
\resultdef{pInjobMultiHeldoutGapPglm}{0.352}
\resultdef{pInjobMultiTrainGapEsmc}{0.357}
\resultdef{pInjobMultiTrainGapMimic}{0.407}
\resultdef{pInjobMultiTrainGapPglm}{0.412}
\resultdef{pInjobRelHeldoutGapEsmc}{0.207}
\resultdef{pInjobRelHeldoutGapMimic}{0.349}
\resultdef{pInjobRelHeldoutGapPglm}{0.225}
\resultdef{pInjobRelHeldoutN}{80}
\resultdef{pInjobRelMixedGapEsmc}{0.238}
\resultdef{pInjobRelMixedGapMimic}{0.466}
\resultdef{pInjobRelMixedGapPglm}{0.460}
\resultdef{pInjobRelMixedN}{38}
\resultdef{pInjobRelvHeldoutGapEsmc}{0.279}
\resultdef{pInjobRelvHeldoutGapMimic}{0.372}
\resultdef{pInjobRelvHeldoutGapPglm}{0.274}
\resultdef{pInjobRelvHeldoutN}{48}
\resultdef{pInjobRelvMixedGapEsmc}{0.368}
\resultdef{pInjobRelvMixedGapMimic}{0.454}
\resultdef{pInjobRelvMixedGapPglm}{0.579}
\resultdef{pInjobRelvMixedN}{30}
\resultdef{pfmLocalizationMax}{0.853}
\resultdef{pfmLocalizationMin}{0.833}
\resultdef{protLengthAccNameMimic}{73.5\%}
\resultdef{protLengthAccOwnMimic}{92.1\%}
\resultdef{protLengthAnnotNameReader}{0.816}
\resultdef{protLengthAnnotReader}{0.802}
\resultdef{protLengthDeltaEsmc}{-0.002}
\resultdef{protLengthDeltaHiEsmc}{0.018}
\resultdef{protLengthDeltaHiMimic}{0.115}
\resultdef{protLengthDeltaHiPglm}{-0.011}
\resultdef{protLengthDeltaLoEsmc}{-0.023}
\resultdef{protLengthDeltaLoMimic}{0.086}
\resultdef{protLengthDeltaLoPglm}{-0.055}
\resultdef{protLengthDeltaMimic}{+0.101}
\resultdef{protLengthDeltaPglm}{-0.033}
\resultdef{protLengthN}{2,036}
\resultdef{protLengthNameEsmc}{0.880}
\resultdef{protLengthNameMimic}{0.880}
\resultdef{protLengthNamePglm}{0.879}
\resultdef{protLengthOwnEsmc}{0.878}
\resultdef{protLengthOwnMimic}{0.981}
\resultdef{protLengthOwnPglm}{0.847}
\resultdef{protLengthPos}{1,187}
\resultdef{proteinTransferGroupMax}{0.961}
\resultdef{proteinTransferGroupMin}{0.754}
\resultdef{raEsmcNameDelta}{+0.041}
\resultdef{raEsmcNameDeltaHi}{+0.068}
\resultdef{raEsmcNameDeltaLo}{+0.013}
\resultdef{raEsmcNameFree}{0.914}
\resultdef{raEsmcNameInstant}{0.873}
\resultdef{raEsmcOwnDelta}{-0.102}
\resultdef{raEsmcOwnDeltaHi}{-0.055}
\resultdef{raEsmcOwnDeltaLo}{-0.148}
\resultdef{raEsmcOwnFree}{0.777}
\resultdef{raEsmcOwnInstant}{0.879}
\resultdef{raMimicNameDelta}{+0.023}
\resultdef{raMimicNameDeltaHi}{+0.053}
\resultdef{raMimicNameDeltaLo}{-0.011}
\resultdef{raMimicNameFree}{0.897}
\resultdef{raMimicNameInstant}{0.874}
\resultdef{raMimicOwnDelta}{-0.122}
\resultdef{raMimicOwnDeltaHi}{-0.069}
\resultdef{raMimicOwnDeltaLo}{-0.177}
\resultdef{raMimicOwnFree}{0.744}
\resultdef{raMimicOwnInstant}{0.866}
\resultdef{raPglmNameDelta}{+0.040}
\resultdef{raPglmNameDeltaHi}{+0.064}
\resultdef{raPglmNameDeltaLo}{+0.016}
\resultdef{raPglmNameFree}{0.915}
\resultdef{raPglmNameInstant}{0.875}
\resultdef{raPglmOwnDelta}{-0.109}
\resultdef{raPglmOwnDeltaHi}{-0.062}
\resultdef{raPglmOwnDeltaLo}{-0.159}
\resultdef{raPglmOwnFree}{0.773}
\resultdef{raPglmOwnInstant}{0.882}
\resultdef{raReportLengthAcc}{88.0}
\resultdef{raReportLengthN}{200}
\resultdef{raReportLengthParsed}{100.0}
\resultdef{raReportLineageAcc}{92.4}
\resultdef{raReportLineageN}{1000}
\resultdef{raReportLineageParsed}{99.3}
\resultdef{raXeightLengthNameDelta}{+0.047}
\resultdef{raXeightLengthNameDeltaHi}{+0.083}
\resultdef{raXeightLengthNameDeltaLo}{+0.010}
\resultdef{raXeightLengthNameFree}{0.944}
\resultdef{raXeightLengthNameInstant}{0.897}
\resultdef{raXeightLengthOwnDelta}{-0.296}
\resultdef{raXeightLengthOwnDeltaHi}{-0.220}
\resultdef{raXeightLengthOwnDeltaLo}{-0.372}
\resultdef{raXeightLengthOwnFree}{0.674}
\resultdef{raXeightLengthOwnInstant}{0.970}
\resultdef{raXeightLineageNameDelta}{+0.018}
\resultdef{raXeightLineageNameDeltaHi}{+0.037}
\resultdef{raXeightLineageNameDeltaLo}{-0.001}
\resultdef{raXeightLineageNameFree}{0.960}
\resultdef{raXeightLineageNameInstant}{0.942}
\resultdef{raXeightLineageOwnDelta}{-0.163}
\resultdef{raXeightLineageOwnDeltaHi}{-0.122}
\resultdef{raXeightLineageOwnDeltaLo}{-0.210}
\resultdef{raXeightLineageOwnFree}{0.801}
\resultdef{raXeightLineageOwnInstant}{0.964}
\resultdef{segmentCompoundFamily}{0.1224}
\resultdef{segmentCompoundGroup}{0.1335}
\resultdef{segmentCompoundParent}{0.5960}
\resultdef{segmentCompoundRole}{0.2918}
\resultdef{segmentProteinCatalytic}{0.2482}
\resultdef{segmentProteinPathway}{0.2898}
\resultdef{trainingMimicMeanFitted}{0.936}
\resultdef{trainingMimicMeanInitial}{0.844}
\resultdef{xdIutP}{0.058}
\resultdef{xdMaxScoreChange}{1.018}
\resultdef{xdOwn}{0.586}
\resultdef{xdOwnHi}{0.670}
\resultdef{xdOwnLo}{0.502}
\resultdef{xdRangeHi}{0.641}
\resultdef{xdRangeLo}{0.546}
\resultdef{xencChoiceAccDrop}{-0.156}
\resultdef{xencChoiceAccDropHi}{-0.108}
\resultdef{xencChoiceAccDropLo}{-0.206}
\resultdef{xencChoiceAccMixed}{0.700}
\resultdef{xencChoiceAccSame}{0.856}
\resultdef{xencChoiceAucDrop}{-0.089}
\resultdef{xencChoiceAucDropHi}{-0.064}
\resultdef{xencChoiceAucDropLo}{-0.118}
\resultdef{xencChoiceAucMixed}{0.827}
\resultdef{xencChoiceAucSame}{0.916}
\resultdef{xencChoiceMajority}{0.600}
\resultdef{xencChoiceN}{150}
\resultdef{xencRelStrictMixed}{0.346}
\resultdef{xencRelStrictMixedHi}{0.363}
\resultdef{xencRelStrictMixedLo}{0.330}
\resultdef{xencRelStrictMixedMinusSame}{+0.024}
\resultdef{xencRelStrictMixedMinusSameHi}{+0.031}
\resultdef{xencRelStrictMixedMinusSameLo}{+0.018}
\resultdef{xencRelStrictN}{894}
\resultdef{xencRelStrictSame}{0.322}
\resultdef{xencRelStrictSameHi}{0.339}
\resultdef{xencRelStrictSameLo}{0.305}
\resultdef{xencRelValidMixed}{0.310}
\resultdef{xencRelValidMixedHi}{0.327}
\resultdef{xencRelValidMixedLo}{0.294}
\resultdef{xencRelValidMixedMinusSame}{+0.031}
\resultdef{xencRelValidMixedMinusSameHi}{+0.038}
\resultdef{xencRelValidMixedMinusSameLo}{+0.024}
\resultdef{xencRelValidN}{1,189}
\resultdef{xencRelValidOwnNllMixedMinusSame}{-0.046}
\resultdef{xencRelValidOwnNllMixedMinusSameHi}{-0.041}
\resultdef{xencRelValidOwnNllMixedMinusSameLo}{-0.053}
\resultdef{xencRelValidSame}{0.279}
\resultdef{xencRelValidSameHi}{0.296}
\resultdef{xencRelValidSameLo}{0.262}
\resultdef{xencRelValidShufNllMixedMinusSame}{-0.016}
\resultdef{xencRelValidShufNllMixedMinusSameHi}{-0.011}
\resultdef{xencRelValidShufNllMixedMinusSameLo}{-0.021}
\resultdef{currentKimiRheaOrdinaryUnseenMolecules}{0.57}
\resultdef{currentKimiRheaOrdinaryFamiliarMolecules}{0.59}
\resultdef{currentKimiRheaUnseenMolecules}{0.75}
\resultdef{currentKimiRheaFamiliarMolecules}{0.80}
\resultdef{currentLocalizationMeanExposurePct}{1.4}

%% file: sections/abstract.tex
Large language models (LLMs) are powerful tools for science, and tool calling enables them to act as
autonomous agents that consult specialist models, run calculations and carry out experiments. To guide an
agent's reasoning, however, specialist information must enter the LLM's context. Current-generation
LLMs generally require that information to be translated into text, images, audio or video---a severe
bottleneck between models that natively process rich, high-dimensional representations. We advocate direct
representation exchange: a mindmeld through which an LLM could receive many properties together,
including information it would not think to request. Such an interface should be extensible, making new
specialists easy to add; flexible, allowing their information to support changing questions and tasks; and
compositional, enabling independently connected specialists to work together without joint training.

We introduce the Latent Interconnect Protocol (LIP) to support this exchange. Its connectors are
learned mappings that bring specialist representations directly into the LLM's internal states. Keeping the language model frozen and only training the connectors makes this approach practical as LLMs grow and evolve. Our demonstration connects three protein and three small molecule specialists to three LLMs,
Gemma 3 (4B and 12B parameters) and Kimi K2.6 (1T total, 32B active parameters)~\cite{moonshot2026kimik26}, while keeping all pretrained models frozen. Independently trained connectors provide useful information without changing the LLM or existing connectors, while supporting both familiar and new question types about unseen objects. We also demonstrate zero-shot transfer to cross-domain tasks without joint training. Task performance and compositionality generally improve with LLM capability, with particularly strong gains in zero shot task transfer. The demonstration also exposes limitations of current LLMs: they were not trained to treat
specialist representations as native inputs, and access to information does not ensure its reliable
interpretation, integration or use during reasoning. We call on specialist developers to provide consistent
representation interfaces and, especially, on LLM developers to train models to receive, combine and reason
with these inputs. Together, these efforts could expand the information available to scientific agents while
preserving the flexibility of conventional tools. 
We release data, pretrained connectors, and code for training, evaluation and interactive inference: \url{https://github.com/PolymathicAI/AI-Mindmeld}.

%% file: sections/intro.tex
\section{Introduction}\label{sec:intro}

Large language models (LLMs) are powerful tools for interpreting scientific literature, developing hypotheses and
reasoning about experiments. Tool calling expands their capabilities further, enabling them to act as autonomous
agents that retrieve data, run calculations and interact with experimental systems
\cite{schick2023toolformer,yao2022react}. Such agents have been used to plan chemical syntheses
\cite{bran2024augmenting}, operate laboratory systems \cite{boiko2023autonomous} and answer biomedical questions using
external tools and databases \cite{huang2025biomni}. They can also consult specialist machine-learning models trained
on scientific data, gaining access to predictions and knowledge about proteins, molecules, cells and other systems
\cite{li2026heterogeneous}.

\begin{figure}[t]
  \centering
  \includegraphics[width=\linewidth]{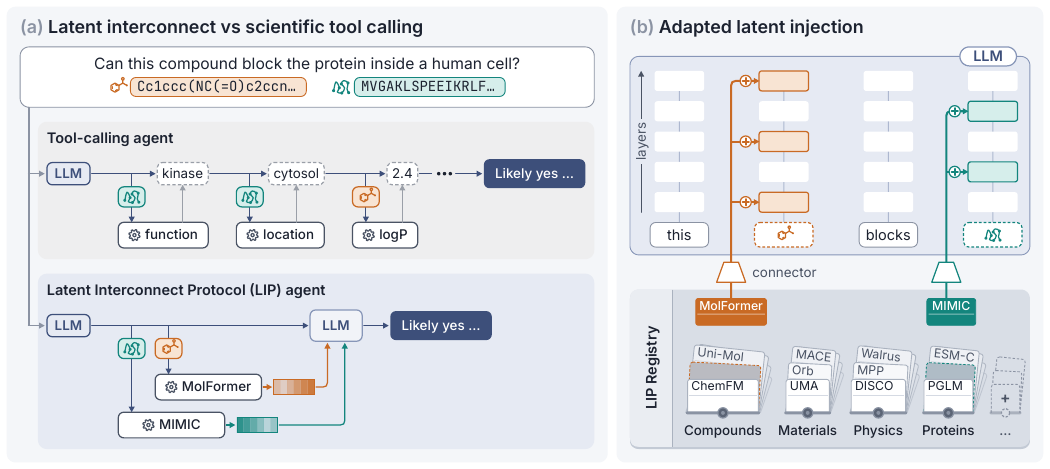}
  \caption{\textbf{Latent Interconnect Protocol alongside tool calling.} \textbf{(a)}~The tool-output route
  returns selected properties as words or numbers; the latent route supplies specialist representations through
  learned connectors. The question and outputs are illustrative, and one agent can use both routes.
  \textbf{(b)}~Connectors selected from the registry write the protein and compound representations into separate
  prompt slots at the marked layers. The $\oplus$ symbols denote the mixing operation defined in
  Section~\ref{sec:framework-connectors}. The layer choices are schematic and may differ between specialists.
  Each representation occupies one token here; LIP also permits multiple-token slots. Registry entries are trained
  independently, and new connectors can be added without changing the shared language model or existing connectors.
  The displayed models and domains illustrate the registry's organization.}
  \label{fig:hero}
\end{figure}

Specialist models may produce many kinds of artifacts, but their contents can inform an LLM's reasoning only if
the LLM can ingest them into its context. For current-generation LLMs, this generally means text, images, video or
audio---the communication media used by humans. Routing information between models through these formats imposes
a severe bandwidth bottleneck, even though both models operate on high-dimensional representations
(Figure~\ref{fig:hero}a). As an example, for a complex query regarding a protein sequence, an agent may need to make explicit queries to specialized tools to calculate or retrieve properties. Making these properties explicit requires selecting them, retrieving
or computing them, and adding the results to the agent's context. Direct representation exchange offers a richer
channel: a \emph{mindmeld} in which an LLM receives specialist information in latent form. One representation can
carry many properties together, including information the LLM does not know or would not think to request.

A framework for this richer exchange should preserve the strengths that make tool use powerful. It should be
\emph{extensible}, making new specialist models easy to add; \emph{flexible}, allowing the same representations to
support different questions and tasks; and \emph{compositional}, enabling information from independently added
specialists to be used together without training on each new combination.
Furthermore, frontier language models embody substantial training investment and continue to evolve through new releases
\cite{kimiteam2025kimik2,moonshot2026kimik26}. Requiring each specialist to adapt and maintain its own LLM would make community-wide extension costly to sustain. A practical interface should therefore keep the
shared LLM frozen and confine alignment to connectors that are economical to fit and refit as models evolve. Together, these properties would allow
a scientific agent to draw on a growing set of models as its questions change.

Prior work connects scientific representations to LLMs \cite{luo2023biomedgpt,tsou2025bioverse,zhuang2024vector},
while the closest framework for independently composed connections focuses on perceptual modalities
\cite{panagopoulou2023x}.
Updating the LLM during alignment compromises independent extensibility by changing the shared interface
\cite{tsou2025bioverse}, and task adaptation through in-context demonstrations is distinct from composing
independently connected specialists without examples of their combination \cite{zhuang2024vector}.
More fundamentally, these interfaces constrain specialist information to the input embedding layer
\cite{luo2023biomedgpt,tsou2025bioverse,zhuang2024vector,panagopoulou2023x}, despite evidence that deeper layers
capture higher-level linguistic information \cite{tenney2019bert} and that multilayer injection improves
performance \cite{meng2024deepstack} (see also Section~\ref{sec:results-ablations-expansion}).
What remains missing is a general framework for specialist models that jointly supports flexibility,
compositionality and extensibility while allowing representations to enter beyond the input embedding layer.

We introduce the Latent Interconnect Protocol (LIP) to implement this form of communication. Under LIP, each
specialist is linked to the LLM by a \emph{connector}: a learned mapping from the specialist's
representation to the LLM's internal states at designated positions in a prompt (Figure~\ref{fig:hero}b).
Our ablations support injecting representations into deeper layers: a single write at a selected internal layer
outperforms a single input write, while injection at six layers gives the highest average performance
(Section~\ref{sec:results-ablations-expansion}). In our
implementation, connectors are trained independently while the specialists and shared LLM remain frozen, so a new
specialist can be added without changing the LLM or existing connectors. A short sentence can then bring
together representations from several specialists. The LLM's existing knowledge and reasoning provide the basis for
using these independently supplied representations together. LIP complements interfaces such as the Model Context
Protocol (MCP) \cite{anthropic2024mcp}; Python execution, database queries and other tools remain indispensable.

Our demonstration connects three protein specialists and three small molecule specialists to three LLMs,
Gemma 3 (4B and 12B parameters) and Kimi K2.6 (1T total, 32B active parameters)~\cite{moonshot2026kimik26}.
These connections support questions about unseen objects and transfer to new question types.
We also test whether independently trained connectors can combine information about the same protein and answer
questions about relationships between proteins and compounds. Task performance and compositionality generally
improve with LLM capability, with particularly strong gains in zero shot task transfer.
Together, these demonstrations show how independently connected specialists can extend an LLM's capabilities
across changing questions and combinations of information sources.

Our implementation illustrates a broader limitation of connecting specialists to current LLMs: the LLM was
not trained to interpret their representations as native inputs. Alignment can make specialist information
accessible without ensuring that the LLM recognizes what kind of input it has received, combines it reliably
with other sources, or uses it consistently during generation. For example, the LLM sometimes describes an
injected protein representation as a familiar word or protein name, and generating a rationale before answering
reduces protein property prediction performance across all three LLMs.

Realizing the broader vision calls for a community effort on both sides of the interface. For LLM developers,
the opportunity is to make specialist representations a native part of model training, explicitly teaching models
to receive, combine and reason with them. Specialist-model developers can support this direction by exposing
representations through consistent, documented interfaces. Shared conventions would make new models easier to
connect without requiring them to learn the same representations. We call on the community to develop these
capabilities so that scientific agents can use the information specialist models already encode, at the bandwidth
of their representations and with the flexibility of ordinary tools.

%% file: sections/related.tex
\subsection*{Related work}\label{sec:related}

We relate LIP to six areas of prior work: text-based model interfaces, LLMs that read specialist representations,
support for multiple domains, support for multiple specialists within a domain, composition without joint training,
and communication through internal states.

\paragraph{Interfaces through text.} Beyond the tool-using agents discussed in the introduction, Socratic Models and
HuggingGPT connect many pretrained models through a language model and exchange information among them as text
\cite{zeng2022socratic,shen2023hugginggpt}. A second family writes biological objects themselves as strings and trains
the language model on them \cite{pei2023biot5,xia2025nature,wang2025txgemma,wang2023instructprotein,levine2024cell2sentence,rizvi2025scaling,bai2025intern}; having
been trained on such mixtures, several of these models already answer questions that relate a protein to a small
molecule in one prompt \cite{pei2023biot5,xia2025nature,wang2025txgemma}. SciCore-Mol and Kamai et al.\ replace the
text interface with a learned representation, but they fine-tune the language model and each serves one domain
\cite{chen2026scicore,kamai2026talking}.

\paragraph{Reading a specialist representation.} Once a representation reaches the language model, it can be read and
described in words. MetaLM proposed language models as a general-purpose interface to other foundation models,
training the LLM together with its connections \cite{hao2022metalm}. Later systems describe images, proteins, molecules, cells and stars from their embeddings
\cite{liu2023visual,carrami2024pqa,luo2023biomedgpt,schaefer2025multimodal,kamai2026talking,abdine2023prot2text,liu2024prott3,zhuo2024protllm,liu2023molca,li2024towards,cao2023instructmol}. Most of them fine-tune the
language model, which can degrade its performance on text-only tasks \cite{zhang2024wings}; those that keep it frozen
typically either insert new modules \cite{alayrac2022flamingo,zhou2025decoding} or supply the representation only at the
input \cite{tsimpoukelli2021multimodal,perozzi2024graph,cheng2024xrag,jin2023time}, as in BLIP-2 and ChatNT \cite{li2023blip,almeida2025multimodal}. DeepStack and ImageBind-LLM inject visual features
at multiple layers while also adapting the LLM \cite{meng2024deepstack,han2023imagebindllm}.
Patchscopes and activation addition write vectors into the residual stream of an unmodified model to interpret or steer it
\cite{ghandeharioun2024patchscopes,turner2023actadd};
we use such a write as a trained interface at several layers of an unmodified model.

\paragraph{Several types of object.} A language model that reads one specialist can in principle read several.
PaLM-E and ChatBridge train on combinations of inputs \cite{driess2023palm,zhao2023chatbridge}.
OneLLM aligns eight modalities through shared components, revisits earlier modalities' data when adding new ones,
and fine-tunes the LLM \cite{han2023onellm}.
Scientific systems that align several types separately, such as BioMedGPT-10B and BioVERSE, evaluate one type per prompt
\cite{luo2023biomedgpt,tsou2025bioverse}, and the general method Vector-ICL connects encoders from many domains but uses
one encoder per task \cite{zhuang2024vector}.

\paragraph{Several models of one type.} The question of breadth has a counterpart within a single type of object, where
different encoders capture different aspects of the same protein or molecule. Existing systems fuse several such
encoders into one jointly trained input \cite{wang2024protchatgpt,liu2024evollama,kim2025mol,xiao2024proteingpt,abdine2023prot2text,kar2024brave}, or offer alternative
encoders that are used one at a time, either with the language model fine-tuned for each \cite{fallahpour2025bioreason}
or through separate projectors into one frozen language model \cite{zhuang2024vector}. We place separately aligned
encoders side by side in separate slots of one prompt.

\paragraph{Composition without joint training.} Placing several separately aligned representations in one prompt is
useful only if the language model can combine them. That separately aligned inputs can be combined without joint training
has been reported for perceptual modalities in a frozen language model by X-InstructBLIP \cite{panagopoulou2023x}.
ImageBind-LLM transfers an image-trained connection to other modalities using their prealigned ImageBind space
\cite{han2023imagebindllm}. Outside language models, ImageBind aligns modalities through a common hub modality
\cite{girdhar2023imagebind}, and BioBridge learns knowledge-graph-supervised transformations between biomedical
models for retrieval \cite{wang2023biobridge}. We evaluate independently
aligned scientific inputs by combining representations of the same protein and answering protein--compound
relationship questions, distinguishing assembly from useful joint use (Section~\ref{sec:results-within}).

\paragraph{Communication through internal states.} More broadly, several studies let models communicate through internal
representations instead of text \cite{chen2026see,rossi2026latent,liu2026beyond}. Language models have exchanged output embeddings \cite{pham2023models}, intermediate
states in Communicating Activations \cite{ramesh2025communicating}, key--value caches in Cache-to-Cache \cite{fu2025cache} and hidden states
\cite{zou2025latent}; CALM composes two frozen language models through cross-attention trained on examples of the combined skill \cite{bansal2024calm}; the
Vision Wormhole connects vision-language models through their visual input pathway \cite{liu2026vision}; and frozen
encoders have been connected through their latent spaces \cite{moschella2023relative,maiorca2023latent}. Two of these
studies argue explicitly that decoding an internal state into text discards information
\cite{ramesh2025communicating,fu2025cache}, a related motivation for latent communication. Our emphasis is the bandwidth cost of selecting and rendering
scientific information as explicit outputs. We develop a direct channel for specialist models that have no language interface, with a language model as the medium through which they communicate.

%% file: sections/framework.tex
\section{Latent Interconnect Protocol}\label{sec:framework}

The mindmeld outlined above calls for \emph{extensibility}: new specialists should be easy to add;
\emph{flexibility}: their representations should support changing questions and tasks; and
\emph{compositionality}: independently connected specialists should work together without training on each new
combination. These desiderata motivate a framework in which each specialist has its own connection to the LLM,
task instructions remain separate from the representation interface, and multiple connections can be assembled
in one prompt. The Latent Interconnect Protocol (LIP) provides one realization of this idea. Here, connectors are
learned independently against a shared frozen LLM, using affine maps to write
specialist representations into its internal states.

\resultfig{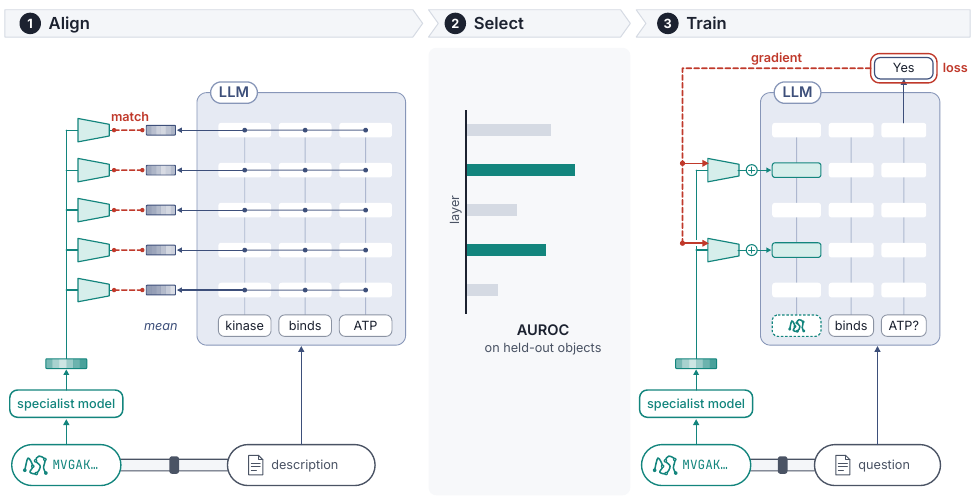}{\textbf{Three stages of independent connector training.}
\textbf{1. Align.} Layer-specific maps match specialist representations to text-derived reference states in the LLM.
The schematic illustrates alignment to description states; the connectors evaluated here instead align to
states obtained from object names.
\textbf{2. Select.} Injection layers are selected to span the LLM's depth, guided by local peaks in single-layer AUROC on held-out scan objects.
\textbf{3. Train.} The selected maps write into the object's slot and are refined by answer loss. Gradients pass through
the frozen LLM to the connector; both pretrained models remain frozen throughout.
Layer counts, bar lengths and example text are illustrative, not measured results or recorded model responses.
Table~\ref{tab:llm_writes} gives the actual schedules.}

A connector binds a particular specialist to this interface: it records the representation to extract, the maps
into the language model's states, and where and how those maps are applied. Figure~\ref{fig:hero}b illustrates a prompt assembled from
such connections. Tools can invoke the specialists; LIP determines how their representations enter the LLM.
Connectors can differ in architecture, injection layers and training procedure. Here, we describe one approach to
constructing them, followed by how LIP registers and combines independently constructed connectors.
Section~\ref{sec:setup} describes the connections constructed for six specialists and three LLMs.
Figure~\ref{fig:training} summarizes the construction procedure used here.

\subsection{Connecting a specialist}\label{sec:framework-connectors}\label{sec:framework-contrast}

A representation enters through designated placeholder tokens, which we call its slot. In Figure~\ref{fig:hero}b,
the protein and compound each receive a slot; the same object could also occupy several slots supplied by different
specialists. LIP allows a slot to span multiple tokens. Our main experiments use one token per representation,
with multiple tokens examined in Section~\ref{sec:results-ablations-expansion}.
The rest of the prompt supplies ordinary language, as in ``Tell me about the protein~$\langle\mathrm{slot}\rangle$.''
The same connector can be reused with different questions in the surrounding text.
The connector writes at the slot, and subsequent tokens read its state through the language model's existing
attention.

We implement the writes as follows. For a specialist embedding $e\in\mathbf{R}^{d_f}$, fixed coordinate-wise standardization gives
$z=(e-\mu)/s$. At each selected layer $\ell$, an affine map produces
$\delta_\ell=W_\ell z+b_\ell\in\mathbf{R}^{d}$, where $d$ is the language model's residual width. These maps align
the specialist's representation with states usable by the LLM. Each connector specifies its representation,
standardization and injection layers.

At each injection point marked by $\oplus$ in Figure~\ref{fig:hero}b, the projected vector is mixed with the slot's
residual state $h_\ell$ at the input of layer $\ell$:
\begin{equation}\label{eq:mix}
  h_\ell \;\leftarrow\; \lVert h_\ell \rVert \, \frac{u_\ell}{\lVert u_\ell \rVert},
  \qquad
  u_\ell = (1-\alpha_\ell)\,\frac{h_\ell}{\lVert h_\ell \rVert}
           +\alpha_\ell\,\frac{\delta_\ell}{\lVert\delta_\ell\rVert}.
\end{equation}
The mixing weight $\alpha_\ell\in[0,1]$ interpolates between the two directions while preserving the incoming norm;
the mixing weights at all selected layers are learned during the training stage.
Only the slot is modified directly; its contents affect subsequent computation through the model's normal
processing. Our implementation supplies a learned, object-independent constant at the token embedding.
Numerical implementation details are given in Supplementary Section~\ref{sec:supp-write-operation}.
On prompts without injected slots, the connector is inactive and the language model's computation is unchanged.

\subsection{Constructing an independent connector}\label{sec:framework-training}\label{sec:framework-layers}

We describe one approach to constructing a connector through alignment, injection layer selection and task training
(Figure~\ref{fig:training}). Both pretrained models remain frozen throughout.
Section~\ref{sec:setup} gives the specific choices used in our experiments.

We start from a dataset of objects paired with textual descriptions and a set of tasks with a metric for evaluating
the LLM's responses. For example, Yes/No questions can be evaluated by AUROC using scores derived from the LLM's
Yes and No token probabilities. The paired descriptions provide alignment targets, while the tasks provide feedback
for layer selection and training.

\textbf{1. Align.} Alignment provides an initial mapping from specialist representations to states the LLM can use
(Figure~\ref{fig:training}, stage 1).
Our approach fits these maps to reference states produced when the LLM processes the paired descriptions.
Because different LLM layers carry different information, each candidate injection layer receives its own map.
This provides a starting point for subsequent task training.

\textbf{2. Select.} Writing at every layer can require billions of connector parameters. A practical connector therefore
needs a limited set of injection layers that makes effective use of the specialist representation.
Evaluating the aligned maps individually on these tasks using the chosen metric provides a performance profile
across the LLM's depth (Figure~\ref{fig:training}, stage 2).
This profile can guide selection toward local performance peaks while retaining coverage of different depths,
allowing the representation to enter at several stages of computation.

\textbf{3. Train.} Each connector is trained independently on these tasks to improve the LLM's use of the specialist's
representations (Figure~\ref{fig:training}, stage 3). Gradients from the task loss pass through the frozen LLM to update
the connector's parameters while both pretrained models remain unchanged. Because trainable parameters and optimizer
state are confined to the connector, adding a specialist does not require a separately adapted copy of the LLM:
the same frozen LLM serves every connector registered for that version.

\subsection{Registry and composition}\label{sec:framework-registry}

The registry in Figure~\ref{fig:hero}b collects connectors for the shared language model. A registry entry identifies
a specialist model and LLM pair, the representation and standardization, and the learned maps and injection schedule. It contains what
is needed to apply the connector without its training data. An entry can be added or updated by fitting only its
connector, leaving the language model and other connectors unchanged. Application details are described in
Supplementary Section~\ref{sec:supp-interface}.

At inference, each slot selects its connector independently. Several slots can describe the same object through
different specialists, or different objects from the same or different domains. The connectors write to their
assigned positions, and the language model's existing knowledge, attention and reasoning provide the means to
combine their information. Sharing the LLM and slot interface allows these combinations to be assembled
without further fitting.
Their usefulness depends on whether the language model can use them together, which motivates the experiments that
follow.

%% file: sections/setup.tex
\resultfig{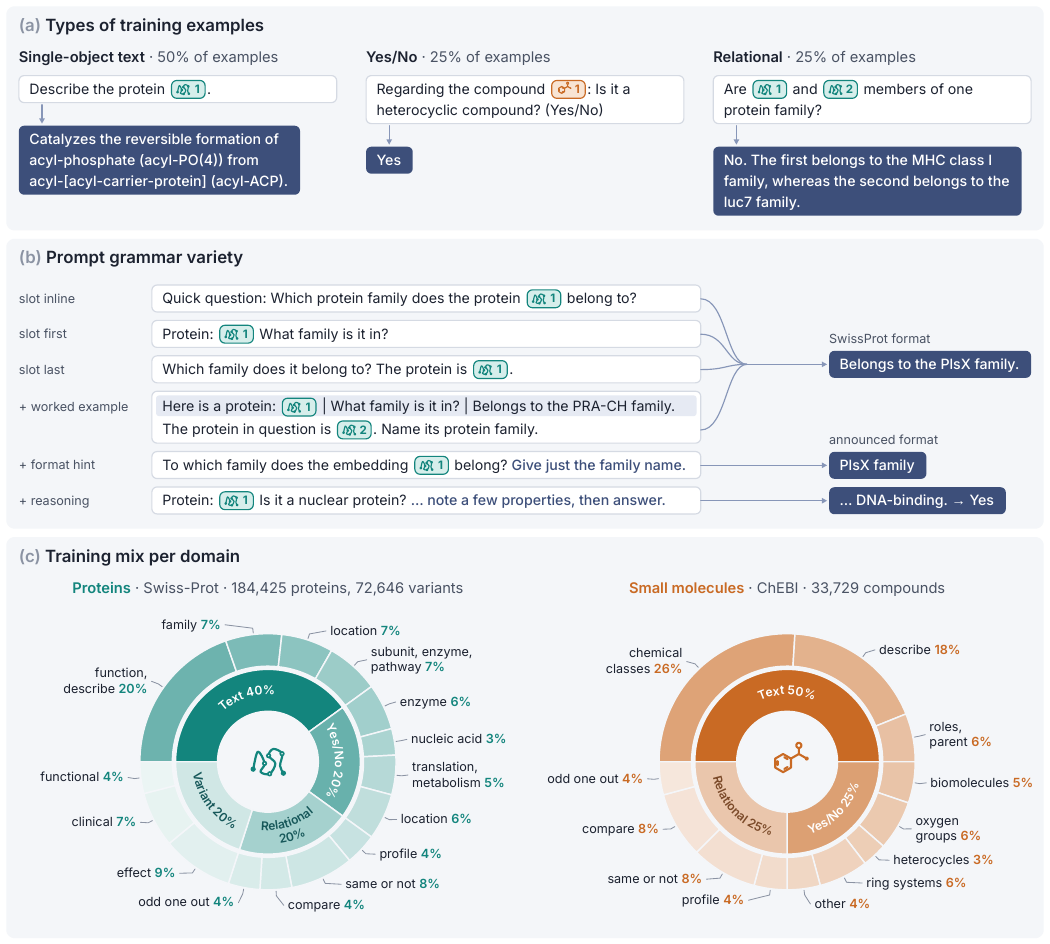}{\textbf{Varied supervision through one connector.}
\textbf{(a)}~Configured shares among non-variant examples. Illustrative prompt--target combinations, not logged draws or generated
answers; the description answer is abbreviated and the relational pair is schematic. The full description field
is truncated at 48 tokens during training. \textbf{(b)}~Prompt variations in slot position, worked context,
answer format and reasoning. All numbered slots within a prompt use the same connector.
\textbf{(c)}~Inner rings show the mix of training streams; outer rings show the question types.
Protein training additionally includes questions about mutation effects, pathogenicity and loss of function.
Percentages refer to the full training mix in each domain. Headers give training-pool sizes;
sampling details are in Supplementary Section~\ref{sec:supp-sampling}.}

\section{Constructing the demonstration}\label{sec:setup}

We demonstrate LIP by connecting protein and small molecule specialists to Gemma 3 (4B and 12B parameters)
and Kimi K2.6 (1T total, 32B active parameters)~\cite{moonshot2026kimik26}. These domains
provide a concrete example of the broader opportunity for latent communication: specialist representations contain
information relevant to many questions, and relationships between their objects create meaningful opportunities
to combine that information. Questions about individual proteins or compounds test the reuse of each connection,
while protein--compound questions test whether independently connected sources can work together.

For each LLM, we connect the same three specialists from each domain, fitting each connector independently while
keeping all pretrained models frozen. This lets us examine reuse across independently learned representations
and how it varies with LLM capability.

We evaluate extensibility through independently added specialists, flexibility through familiar and new questions
about unseen objects, and compositionality through joint questions using independently trained connectors within
and across domains. These evaluations also reveal how the ability to interpret and combine latent inputs varies
with LLM capability, identifying what must improve to realize the broader vision.

\subsection{Models and representations}\label{sec:setup-models}

The protein specialists are MIMIC~\cite{golkar2026mimic},
ESM-C~\cite{candido2026language} and PGLM~\cite{chen2025xtrimopglm}, each receiving an amino-acid sequence.
The molecule specialists are MoLFormer~\cite{ross2022large}, Uni-Mol2~\cite{ji2024unimol2} and
ChemFM~\cite{cai2025chemfm}.

For each specialist, we inspect read layer performance using property probes and, for proteins, variant probes.
Based on these, we select up to two layers with complementary performance, retaining their concatenation when it improves
performance on at least one task. These choices precede alignment, and the same selected representations are used
across all three LLMs. Table~\ref{tab:connector_instances} lists the representations and their dimensions.
Read probes and layer profiles are given in Supplementary Section~\ref{sec:supp-read-probes}.

\begin{table}[htbp]
\centering
\setlength{\abovecaptionskip}{5pt}
\input{tables/specialist_reads}
\caption{\textbf{Specialist read layers and representations.}
Read layers were chosen using property probes and, for proteins, variant probes; complementary layers were
concatenated when they improved performance on at least one task.}
\label{tab:connector_instances}
\end{table}

Each representation occupies one token in the main experiments. Full model and extraction details are given in
Supplementary Section~\ref{sec:supp-models}.

\subsection{Prompts, data and evaluation}\label{sec:data}\label{sec:setup-training}\label{sec:setup-evaluation}

\paragraph{Prompt construction.}
We use varied supervision so that each connector is trained to make an object's representation useful for different
questions. Single-object text targets ask for descriptive information, Yes/No targets focus on individual properties,
and relational targets ask about multiple objects (Figure~\ref{fig:data_figure}a). Reference answers are constructed
from the corresponding annotations. Both domains use these three example types, drawing on protein function, family and location
annotations or compound definitions, classes and relationships. Protein training additionally includes questions
about mutation effects, pathogenicity and loss of function; Figure~\ref{fig:data_figure}c shows the training mix for
each domain. All slots in relational prompts use the same connector, preserving independent
connector training; combinations of different connectors are reserved for evaluation.

To encourage reuse across ways of asking a question, a compositional prompt grammar augments text and Yes/No
examples by varying wording, object references, slot placement, requested answer format and reasoning. These variations are
intended to make the representation usable across linguistic contexts while allowing the surrounding instruction
to determine the response. Figure~\ref{fig:data_figure}b illustrates variations in slot position, worked context,
answer format and requests for reasoning. Worked examples use other training objects, with every slot
supplied through the same connector. Supervision categories and sampling rules are in Supplementary
Section~\ref{sec:supp-data}, with question examples in Section~\ref{sec:supp-prompts}.

\resultfig{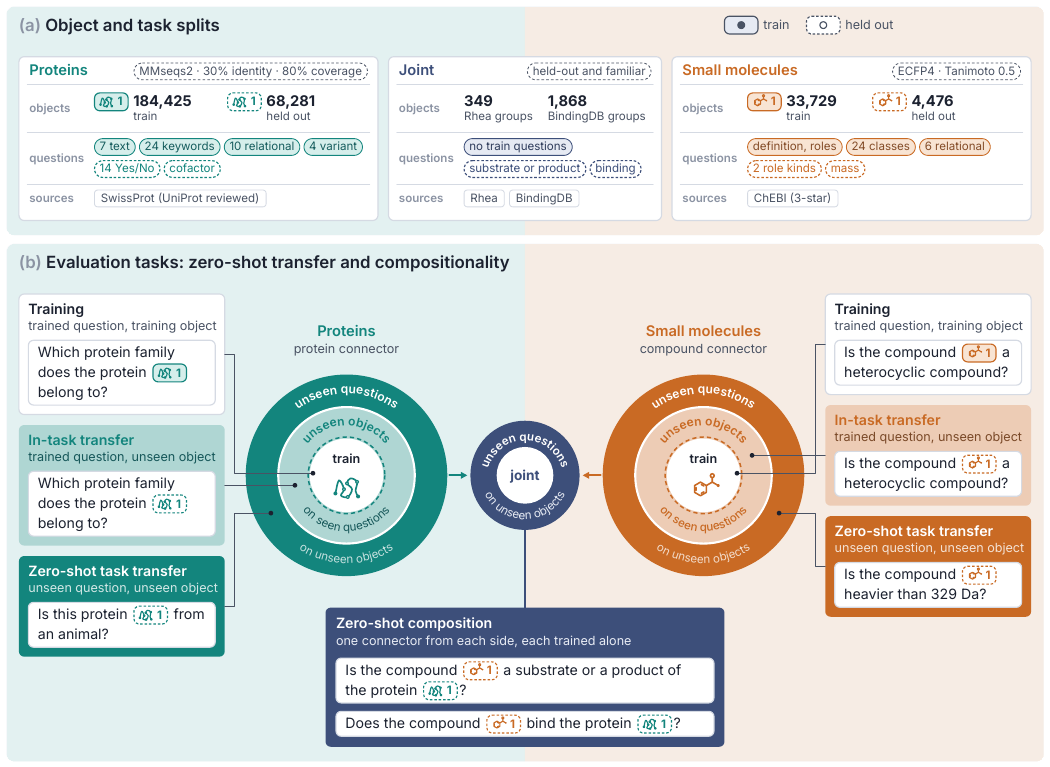}{\textbf{Object, task and cross-domain evaluation design.}
\textbf{(a)}~Training and held-out object pools, question types and data sources. Training counts give the pools
used by the fitted connectors; held-out counts give the full validation and test splits, not evaluated sample sizes.
The joint card reports evaluation groups from Rhea and BindingDB for reaction participation and protein--compound
binding. Cross-domain evaluations include both familiar and held-out objects.
\textbf{(b)}~The progression from trained questions about held-out objects to held-out questions about held-out
objects, then cross-domain questions using independently trained connectors. Solid slots denote training objects
and dashed slots denote held-out objects. The example questions are illustrative. Within-domain composition of
different connectors is an additional evaluation described in the text. Final task definitions, scored populations
and controls are in Supplementary Section~\ref{sec:supp-evaluation}.}

\paragraph{Data split.}
To evaluate generalization beyond the objects used to fit a connector, we separate training and held-out pools
before constructing examples (Figure~\ref{fig:eval_figure}a). Protein sequences and annotations come from
Swiss-Prot, the reviewed part of UniProt~\cite{uniprot2025uniprot}; compounds come from the three-star entries of
ChEBI~\cite{malik2026chebi}. The molecule training pool is substantially smaller, with approximately
34,000 compounds compared with 184,000 proteins, and covers a narrower range of annotation types;
protein supervision additionally includes variant effects (Figure~\ref{fig:data_figure}). The protein split uses an MMseqs2 exclusion criterion of 30\% sequence identity and
80\% coverage~\cite{steinegger2017mmseqs2}; molecules are separated at ECFP4 fingerprint Tanimoto similarity 0.5
after grouping related chemical forms. The exclusion applies to every object in a connector-training prompt,
including worked examples and relational partners, so those contexts cannot reintroduce held-out objects.

Layer selection uses scan objects from the training pool, excluded from fitting the maps
used in the scans. Split construction is detailed in Supplementary Section~\ref{sec:supp-splits}.
These exclusions concern connector training, not exposure during the LLM or specialist model's pretraining.

\paragraph{Alignment and injection layer selection.}
We implement the \emph{Align} and \emph{Select} stages outlined in
Section~\ref{sec:framework-training} as follows. We fit ridge maps from the selected specialist representations to LLM states at object names
in prompts containing a description followed by the name. We average the name token states for proteins and use
the final name token state for molecules. Names are needed for these fitting objects, but not when using the
trained connector. Name and description reference targets are compared in Supplementary Section~\ref{sec:supp-reference-states}.

We evaluate each aligned map separately on the scan objects using five diagnostic Yes/No
questions per domain. Mean task AUROC gives a profile across injection layers for each connector; injecting the
LLM's own name token states provides a reference profile. For each domain and LLM, we take the maximum across
these profiles at each layer to form an envelope. Guided by its local maxima, we select layers spanning the LLM's
depth and use the resulting schedule for all specialists in that domain. Table~\ref{tab:llm_writes} lists the schedules,
with profiles shown in Supplementary Figure~\ref{fig:supp_write_profiles}.

\begin{table}[htbp]
\centering
\setlength{\abovecaptionskip}{5pt}
\input{tables/llm_writes}
\caption{\textbf{LLM injection layers.}
Shared layers for each domain and LLM were chosen to span its depth, guided by local peaks in the maximum AUROC
envelope over connectors and the LLM's own-state reference.}
\label{tab:llm_writes}
\end{table}

At every selected injection layer, each connector's mixing weight is initialized from its mean task AUROC
above chance: we subtract 0.5, normalize by the best layer's above-chance score and clip to $[0.1, 0.9]$.
All these weights are then learned during training. Reference layer choices, diagnostic questions and fitting
details are given in Supplementary Sections~\ref{sec:supp-ridge-fitting} and~\ref{sec:supp-write-selection}.

\paragraph{Training and evaluation tasks.}
We deliberately withhold selected task categories to test whether a trained connection can support new questions
(Figure~\ref{fig:eval_figure}a). Task training uses the varied annotation-derived questions above, with no object-naming response target.
Reserved tasks include protein lineage and length, selected compound roles and compound mass. This restricted
supervision creates a test of zero-shot transfer; applications prioritizing performance on known tasks could train
on a broader range of relevant questions, including categories withheld here. Selected grammar entries are also
reserved to test new phrasings. Withholding a question does not remove every related clue from the remaining
training annotations.

The evaluations distinguish three forms of reuse (Figure~\ref{fig:eval_figure}b). Trained questions on unseen
objects test generalization beyond the fitting pool; new questions on unseen objects test flexibility beyond
supervised tasks; and joint questions test composition of independently trained connections. Within-domain
comparisons combine different protein connectors. Cross-domain comparisons use Rhea reaction-participation and BindingDB protein--compound
binding questions, bringing protein and compound connectors together without any cross-domain training stream.
These evaluations include both familiar and unseen objects, with the available combinations determined by task
and filtering. The Rhea comparison in Figure~\ref{fig:results_merging}c contains only unseen proteins: nearly every
seen-protein pair was excluded because its relation was already stated in the protein connector's training descriptions.
Different connectors are combined only at evaluation, without further
fitting. We separately assess how response mode affects use of the supplied information. Main comparisons use
the final 2,000-step checkpoint of each connector.

\subsection{Training implementation and cost}\label{sec:setup-cost}

We train each connector with a target batch size of 85 examples per optimizer step. Answer-token losses are
weighted by answer type, semantic segment and position within each segment, emphasizing informative content over
predictable connecting text, and normalized by the total token weight. We use AdamW with learning rates for the
maps, biases and input constant scaled to their initial magnitudes, a 20-step linear warmup, and a constant rate
thereafter. Following the Trellis optimization recipe~\cite{foote2026trellis}, we distribute expert computation
across eight GPUs and pack examples to avoid processing padding, leading to $5.1\times$ and $1.19\times$ improvements in training-step
throughput, respectively.

Alignment and layer selection took tens of minutes for Gemma and a few hours for Kimi K2.6, using precomputed
specialist representations. Training for 2,000 updates took roughly three hours on one H100 for Gemma 3 4B,
four hours on one H200 for Gemma 3 12B, and 12 hours on eight H200s for Kimi K2.6 per connector.
Detailed training settings are given in Supplementary Section~\ref{sec:supp-training}.

%% file: tables/specialist_reads.tex
\begingroup
\small
\setlength{\tabcolsep}{5pt}
\renewcommand{\arraystretch}{1.12}
\noindent\begin{tabularx}{\linewidth}{@{}>{\raggedright\arraybackslash}p{0.23\linewidth}Xr@{}}
\toprule
\textbf{Specialist} & \textbf{Representation} & \textbf{Dim.} \\
\midrule
\multicolumn{3}{@{}l}{\color[HTML]{13857D}\textbf{Proteins}}\\[2pt]
MIMIC & Layer 16 (mean pooled) $\Vert$ Layer 20 (mean pooled) & 3,072 \\
ESM-C 600M & Layer 33 (mean pooled) $\Vert$ Layer 36 (mean pooled) & 2,304 \\
PGLM-1B & Layer 24 (mean pooled) & 2,048 \\
\addlinespace[5pt]
\multicolumn{3}{@{}l}{\color[HTML]{C96A24}\textbf{Molecules}}\\[2pt]
MoLFormer-c3 & Layer 9 (mean pooled) $\Vert$ Layer 12 (EOS token) & 1,536 \\
Uni-Mol2 1.1B & Layer 23 (CLS token) $\Vert$ Layer 23 (mean pooled) & 3,072 \\
ChemFM-1B & Layer 17 (last token) $\Vert$ Layer 17 (mean pooled) & 4,096 \\
\bottomrule
\end{tabularx}
\par\vspace{3pt}
\raggedright\footnotesize Dim.: extracted representation width; $\Vert$: concatenation in the listed order.
\par\endgroup

%% file: tables/llm_writes.tex
\begingroup
\small
\setlength{\tabcolsep}{5pt}
\renewcommand{\arraystretch}{1.12}
\noindent\begin{tabular*}{\linewidth}{@{}l@{\extracolsep{\fill}}ll@{}}
\toprule
\textbf{LLM} & {\color[HTML]{13857D}\textbf{Proteins}} & {\color[HTML]{C96A24}\textbf{Molecules}} \\
\midrule
Gemma 3 4B & 2, 7, 11, 17, 26, 33 & 4, 7, 12, 16, 21, 32 \\
Gemma 3 12B & 6, 11, 16, 23, 33, 44 & 4, 9, 17, 23, 31, 41 \\
Kimi K2.6 & 7, 13, 22, 33, 49, 58 & 7, 13, 22, 32, 49, 57 \\
\bottomrule
\end{tabular*}
\par\endgroup

%% file: sections/results.tex
\section{Results}\label{sec:results}

For each of the three LLMs, we train six connectors independently on a subset of objects and tasks, without
exposure to other connectors. We evaluate generalization to unseen objects, zero-shot transfer to new tasks,
and composition within and across domains, examining how these capabilities vary with LLM capability.
We then examine the LLM's responses to latent inputs, the effect of reasoning, and how injection depth and token
count affect performance.

\subsection{Within-task generalization}\label{sec:results-access}

We first evaluate whether connectors can answer familiar types of questions on previously unseen objects. As described in Section~\ref{sec:data}, the proteins
and molecules are excluded from connector fitting, and the questions use wording withheld from training.
For each Yes/No question, we rank objects by the LLM's relative probability of Yes versus No and report the
area under the receiver operating characteristic curve (AUROC), with 0.5 indicating chance discrimination.
Scoring and uncertainty are described in Supplementary Section~\ref{sec:supp-protocol}.

\resultfig{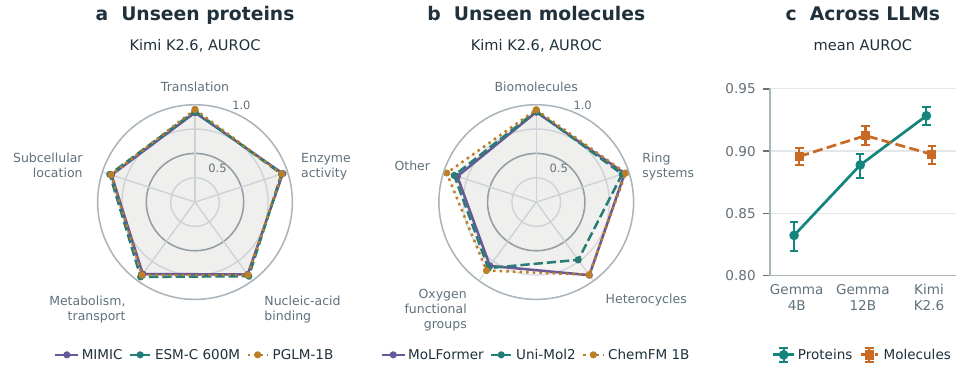}{\textbf{Within-task generalization to unseen objects.}
\textbf{(a--b)}~Task-group AUROC for protein and molecule connectors attached to Kimi K2.6.
Protein connectors perform similarly across groups, while molecule connectors differ most on heterocycles.
\textbf{(c)}~Mean AUROC across the three specialists in each domain for Gemma 3 4B, Gemma 3 12B and Kimi K2.6.
Protein performance increases across these LLMs, whereas molecule performance varies less and peaks with Gemma 3 12B.
Error bars in (c) show 95\% bootstrap confidence intervals.}

With Kimi K2.6, all six connectors support strong performance on objects excluded from
connector fitting. Across the three protein connectors, mean AUROC over trained tasks ranges from
\result{currentProteinAccessMin} to \result{currentProteinAccessMax}, with similar performance across connectors
on questions about enzyme activity, nucleic acid binding, translation, metabolism and transport, and subcellular
location (Figure~\ref{fig:results_access}a). The three molecule connectors achieve mean AUROCs of
\result{currentMoleculeAccessMin}--\result{currentMoleculeAccessMax} across trained tasks, with greater variation
across chemical classes and the largest differences on heterocycles (Figure~\ref{fig:results_access}b).
A matched control for unseen question wording is given in Supplementary Section~\ref{sec:supp-wording}.

Averaging across the three specialists in each domain reveals different trends across LLMs
(Figure~\ref{fig:results_access}c). Protein performance increases from Gemma 3 4B through Gemma 3 12B to
Kimi K2.6. Molecule performance varies less and peaks with Gemma 3 12B. On new tasks, however, Kimi K2.6
substantially outperforms both Gemma models in both domains (Section~\ref{sec:results-transfer}).

\subsection{Zero-shot transfer to new tasks}\label{sec:results-transfer}

We next evaluate new tasks on unseen objects. Beyond explicitly excluding these question types from connector task
training, we filter the free-text descriptions used as training targets to reduce disclosure of information related to these
questions. We then classify a task as zero-shot only when the estimated remaining description exposure is below
0.1\% of training examples (Supplementary Section~\ref{sec:supp-zero-shot-exposure}).

\resultfig{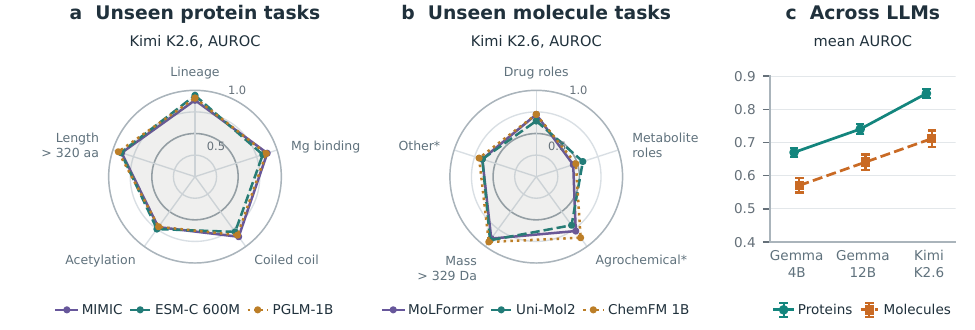}{\textbf{Zero-shot transfer to new tasks on unseen objects.}
\textbf{(a--b)}~Task-group AUROC for protein and molecule connectors attached to Kimi K2.6.
Protein connectors show similar transfer performance, while molecule transfer is strongest for mass.
\textbf{(c)}~Mean AUROC across the three specialists in each domain for Gemma 3 4B, Gemma 3 12B and Kimi K2.6.
Performance increases across these LLMs in both domains.
Error bars in (c) show 95\% bootstrap confidence intervals.
$^{*}$Starred question types had exposure up to 0.3\%, and in some cases related information may have been implied
in the training examples' free-text descriptions.}

With Kimi K2.6, the three protein connectors show strong transfer across questions about lineage, magnesium
binding, coiled-coil regions, acetylation and length (Figure~\ref{fig:results_transfer}a),
with the highest scores for lineage or length.

Molecule transfer is strongest for predicting whether mass exceeds 329 Da, with AUROC
\result{currentMoleculeTransferMassMin}--\result{currentMoleculeTransferMassMax} across the three connectors
(Figure~\ref{fig:results_transfer}b). Transfer to chemical roles is weaker and more variable: drug-role AUROC
ranges from \result{currentMoleculeTransferDrugMin} to \result{currentMoleculeTransferDrugMax}, while
metabolite-role performance is near chance, at \result{currentMoleculeTransferMetaboliteMin}--\result{currentMoleculeTransferMetaboliteMax}.
The figure also reports agrochemical and other roles, including the starred question types with greater
estimated description exposure. Task definitions and populations are given in Supplementary
Section~\ref{sec:supp-transfer-populations}.

Averaging across the three specialists in each domain, transfer performance increases from Gemma 3 4B through
Gemma 3 12B to Kimi K2.6 (Figure~\ref{fig:results_transfer}c). The gains are evident in both domains, including
molecules, for which performance on trained tasks varies much less across LLMs. These results show that the
connections support questions beyond their task supervision, with stronger transfer in the more capable LLMs.

\FloatBarrier
\begingroup
\setlength{\abovecaptionskip}{3pt}
\setkeys{Gin}{trim=0bp 12bp 0bp 0bp,clip}
\resultfig{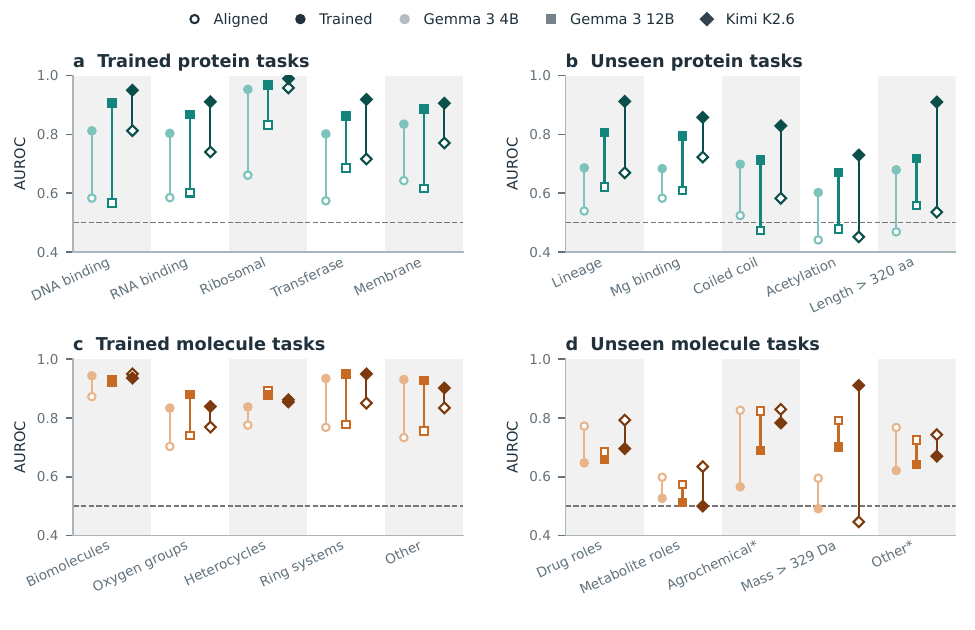}{\textbf{Effects of connector training on trained and unseen tasks.}
Open and filled markers show AUROC after alignment and after connector training, respectively, averaged across
the three specialists in each domain for each LLM.
\textbf{(a--b)}~Training improves performance on both trained and unseen protein tasks.
\textbf{(c)}~Training generally improves performance on trained molecule tasks.
\textbf{(d)}~Performance declines on most unseen molecule tasks; mass prediction improves only for Kimi K2.6.
$^{*}$Starred question types had exposure up to 0.32\%, and in some cases related information may have been implied
in the training examples' free-text descriptions.}
\endgroup

To examine what task training adds beyond alignment, Figure~\ref{fig:results_training_comparison} compares the performance of the
same connectors before and after \emph{Train}
(Section~\ref{sec:framework-training}). Each point averages the three specialists in a domain for one LLM.
Before task training, the aligned representations already support predictions on a range of trained and unseen
questions.

For proteins, training improves performance across the displayed trained and unseen task groups with all three
LLMs (Figure~\ref{fig:results_training_comparison}a--b). Molecule training generally improves performance on trained
tasks, but reduces performance on most unseen tasks (Figure~\ref{fig:results_training_comparison}c--d).
Mass prediction improves only with Kimi K2.6, illustrating that the effect of training also depends on the LLM.

These patterns suggest that connector training can improve general access to specialist information or favor
the supervised properties at the expense of other information. The broad protein gains are consistent with
improved access, while the molecule transfer losses suggest greater specialization toward the training tasks.
Transfer may arise because the frozen LLM can combine accessible object properties with relationships learned
during pretraining to answer new questions. A more capable LLM may be better able to use this information,
consistent with the transfer trend in Figure~\ref{fig:results_transfer}c.

We next ask whether independently learned connections can also be used together.

\subsection{Zero-shot compositionality}\label{sec:results-within}

Compositionality requires independently trained connections to remain useful when brought together.
We test two forms of this capability: combining different specialist views of the same object, and answering
questions about relationships between objects from different domains. Neither combination receives joint training.

For information merging, we supply three representations of the same unseen protein, one each from
MIMIC, ESM-C and PGLM. We compare each connector individually with all three used together, asking trained
and unseen questions with Gemma 3 4B, Gemma 3 12B and Kimi K2.6.
Controls using repeated copies of one specialist are reported in Supplementary Section~\ref{sec:supp-same-object}.

\begingroup
\setlength{\abovecaptionskip}{3pt}
\resultfig{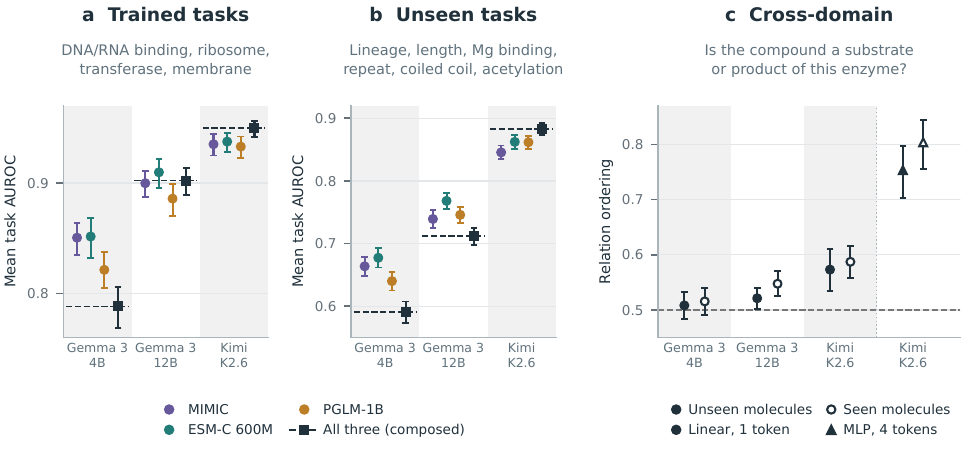}{\textbf{Zero-shot composition of independently trained connectors.}
Compositionality improves with LLM capability, with Kimi K2.6 benefiting most from combining independently trained connectors.
\textbf{(a--b)}~Individual and combined protein connectors on trained and unseen tasks.
\textbf{(c)}~Rhea reaction-participation ordering. The first three columns compare LLMs using ESM-C 600M
and ChemFM 1B with single-token linear connectors. The final column shows Kimi K2.6 using ESM-C 6B and
ChemFM 3B with four-token MLP connectors (Section~\ref{sec:results-ablations-expansion}).
Point estimates are higher for seen molecules.
All proteins are unseen: nearly every seen-protein pair was excluded because the protein connector's training descriptions
already stated the queried relation.
Error bars show 95\% bootstrap confidence intervals.}
\endgroup

The benefit of combining representations depends on the LLM (Figure~\ref{fig:results_merging}a--b).
With Gemma 3 4B, the combination performs worse than every individual connector on both trained and unseen
tasks. With Gemma 3 12B, it approaches the best individual connector on trained tasks but remains below all
three on unseen tasks. With Kimi K2.6, combining the representations improves mean AUROC over every individual
connector on both task groups. Thus, the most capable LLM tested can benefit from multiple specialist
representations without joint connector training, while additional inputs can reduce performance in smaller LLMs.

We next test composition across domains by connecting ESM-C 600M and ChemFM 1B to each LLM. Given a protein
embedding and a compound embedding, can the LLM tell whether the protein catalyzes a reaction in which the
compound is a substrate or product?
We construct this task from UniProt-linked annotations in Rhea, a curated database of biochemical
reactions~\cite{bansal2022rhea}. Each matched group starts with two annotated protein--compound pairs;
swapping their compounds gives two mismatched pairs made from the same objects. We measure how often the LLM
scores an annotated pair above a mismatched pair sharing its protein or compound. Because each object appears
in both types of pair, this tests their relationship rather than a preference for particular objects; chance
ordering is 0.5. Task populations and scoring are detailed in Supplementary Section~\ref{sec:supp-cross-domain}.

All evaluated proteins are unseen. We exclude relationships stated in the protein connector's training
descriptions, which leaves almost no eligible pairs containing familiar proteins. We therefore compare unseen
proteins paired with either familiar or unseen molecules (Figure~\ref{fig:results_merging}c). Both settings test
a new relationship question: the connectors were trained independently, without protein--compound relationship
questions or access to one another's representations.

With the single token linear connectors, ordering performance increases from Gemma 3 4B through Gemma 3 12B
to Kimi K2.6. Gemma 3 4B performs near chance, whereas Kimi reaches approximately
\result{currentKimiRheaOrdinaryUnseenMolecules} with unseen molecules and
\result{currentKimiRheaOrdinaryFamiliarMolecules} with familiar molecules. Using the larger ESM-C 6B and ChemFM 3B
specialists with four token MLP connectors (Section~\ref{sec:results-ablations-expansion}), Kimi reaches
\result{currentKimiRheaUnseenMolecules} and \result{currentKimiRheaFamiliarMolecules}, respectively, showing the
potential of LIP beyond single token linear injections. Familiar molecules have higher point estimates in every
configuration. Together with the protein composition results, this shows that independently trained connectors
can support joint use within and across domains.

\subsection{LLM responses to latent inputs and the effect of reasoning}\label{sec:results-reasoning}

\begingroup
\setlength{\abovecaptionskip}{3pt}
\resultfig{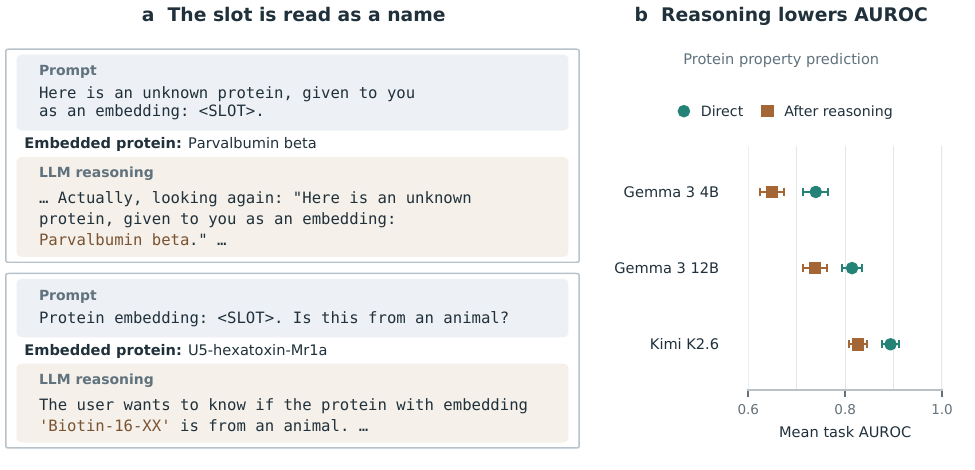}{\textbf{LLM responses to latent inputs and the effect of reasoning.}
\textbf{(a)}~Examples of the LLM treating injected protein embeddings as protein names.
\textbf{(b)}~Mean task AUROC for direct answers and answers after reasoning, averaged across protein connectors for each LLM.
Reasoning reduces protein-property prediction performance for all three LLMs.
Bars in (b) show 95\% object bootstrap confidence intervals, using shared draws across tasks and connectors.}
\endgroup

The preceding results show that the LLM can use injected embeddings to predict properties and relationships.
However, this form of input is out of distribution for the pretrained LLM and can also lead to undesirable behavior. During reasoning with a protein connector, for example, the LLM regularly treats the injected slot as though the user had supplied a protein name~(Figure~\ref{fig:results_slot_name}a). This interpretation is unsurprising because the LLM did not encounter these protein embeddings during its own training. Mapping an embedding to a familiar protein name is itself a nontrivial feat: it connects a specialist model's representation to the extensive biological knowledge available to frontier LLMs.
This raises a central question: does LIP work by mapping an embedding to a protein name and recalling facts
associated with it, or does the embedding provide useful information beyond that name?

We first examine reasoning, during which this interpretation of the embedding as a name becomes explicit.
Averaged across protein connectors, reasoning reduces mean task AUROC for Gemma 3 4B, Gemma 3 12B and
Kimi K2.6 (Figure~\ref{fig:results_slot_name}b). We request brief reasoning with a 200-token generation limit
and then score Yes/No answers, retaining cases that reach the limit. The comparison therefore measures
discrimination under this reasoning procedure, rather than the accuracy of naturally completed answers.
In a separate control, when Kimi K2.6 receives the protein's true name without latent injection, reasoning
increases mean task AUROC by 0.03.

\begingroup
\setlength{\abovecaptionskip}{3pt}
\resultfig{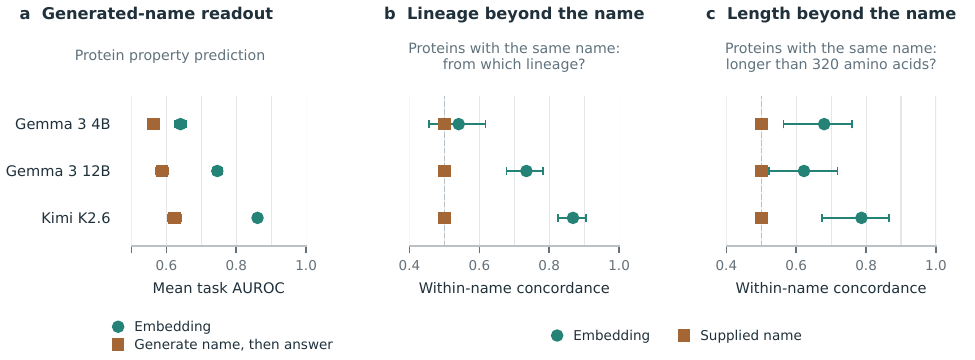}{\textbf{Latent inputs preserve useful distinctions beyond names.}
\textbf{(a)}~Protein-property prediction using embeddings or names generated from those embeddings.
Retaining the embedding gives higher AUROC for all three LLMs.
\textbf{(b--c)}~Lineage and length discrimination among proteins sharing the same name, measured by within-name concordance;
lineage uses four low-exposure questions, and length tests whether a protein exceeds 320 amino acids.
Embeddings provide information beyond the shared name, with Kimi K2.6 showing the strongest performance in both tests.
Supplied names score 0.5 by construction. Error bars show 95\% bootstrap confidence intervals.}
\endgroup

To test whether this name readout accounts for LIP's performance, we ask the LLM to infer a name from the protein embedding, then replace the embedding with that name and ask the same downstream questions. This replacement reduces mean task AUROC for all three LLMs (Figure~\ref{fig:results_reasoning}a).
This provides evidence that the embedding provides useful information beyond what the LLM recalls from the inferred name. However, the loss could reflect incorrect name inference rather than information missing from a correct name.

We therefore next ask whether the embedding provides useful information beyond even the correct supplied name.
Among proteins with identical database-name strings, we test four low-exposure lineage questions and whether
length exceeds 320 amino acids (Figure~\ref{fig:results_reasoning}b--c). A deterministic name-only score must
tie within each group, since both the name and the question are identical. Embeddings yield mean within-name
concordance above 0.5 for both tests with all three LLMs, with Kimi K2.6 performing best in each. These results
show that the latent input supplies useful distinctions beyond the shared name. Name-substitution and
within-name controls are defined in Supplementary Section~\ref{sec:supp-beyond-name}.

Together, these results show that LIP's usefulness cannot be explained solely by recalling information associated
with the supplied object name. One interpretation of the reasoning decline is that the LLM maps the unfamiliar
latent input to a familiar name and then reasons mainly from that name, leaving other useful information underused.
Given the central role of reasoning in modern LLM use, this points to a need for frontier language models trained to work with information-rich latent placeholders and draw on their content throughout the reasoning process.

\FloatBarrier
\subsection{LIP ablations, extensions and scaling}\label{sec:results-ablations-expansion}

Throughout the preceding experiments, each connector represented an object with one token, injected at
multiple LLM layers. The specialists vary in size; when multiple versions were available, we selected the one
closest to one billion parameters. We now examine how injection depth, token count, specialist scale and
connector architecture affect performance, and compare the resulting connectors with supervised linear baselines.

\begingroup
\setlength{\abovecaptionskip}{3pt}
\resultfig{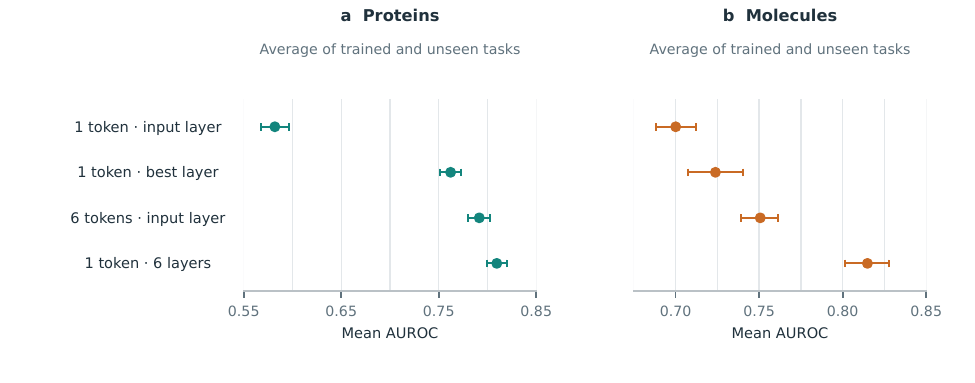}{\textbf{Effects of write location and token count.}
Gemma 3 12B with MIMIC \textbf{(a)} and ChemFM-1B \textbf{(b)}. Mean of trained and unseen task performance; error bars show 95\% bootstrap intervals.
The paper's connector uses one token and six injection layers.
Moving a single-token write from the input to the best layer improves performance, supporting deep injection.
Six input tokens match the six-layer connector's parameter count but require more inference compute;
even so, performance is higher than single-token injection at the best layer but below the six-layer connector.}
\endgroup

\paragraph{Injection depth.}
We examine how injection depth and token count affect performance using Gemma 3 12B with MIMIC for
proteins and ChemFM 1B for molecules (Figure~\ref{fig:results_write_ablation}). In both domains, moving a
single-token write from the input to the best internal layer improves mean AUROC. Writing to the same token
at six selected layers improves performance further, supporting the use of deep, repeated injection in the
connectors used throughout the paper.

To compare repeated injection with a larger input interface, we also supply six tokens at the input layer.
This matches the connector parameter count of one token with six injection layers, but requires more inference
compute because the additional tokens pass through the LLM. Despite this additional compute, six input tokens
achieve higher mean AUROC than one token at the best layer but remain below one token at six layers. These results suggest
that, at matched connector parameter count, distributing injection across layers is more effective than
supplying additional tokens at the input.

\begingroup
\setlength{\abovecaptionskip}{3pt}
\resultfig{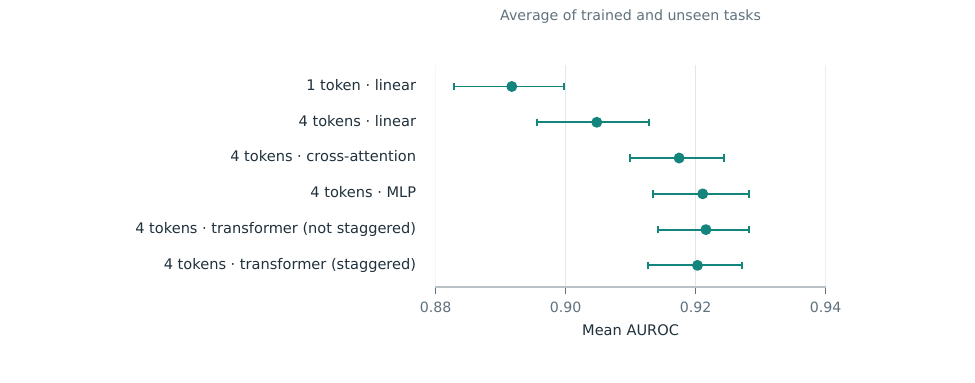}{\textbf{Connector extensions.} ESM-C 6B with Kimi K2.6:
mean AUROC across trained and unseen tasks after 2,000 training steps. Error bars show 95\% bootstrap intervals.}
\endgroup

\paragraph{Beyond single token linear injections.}
LIP permits richer interfaces than a single token with linear injection maps. To explore these extensions
with the aim of maximizing performance, we connect ESM-C 6B to Kimi K2.6 and compare four-token connectors
with linear maps, MLPs and transformer blocks (Figure~\ref{fig:results_extensions}). We also compare using
the same injection layers for all tokens with staggering their injection layers. Mean AUROC across trained
and unseen tasks increases from approximately 0.89 for the single-token connector to 0.90 with four-token
linear maps and 0.92 with MLP or transformer connectors.

The MLP performs comparably to both transformer variants. After 2,000 training steps, the transformer blocks'
attention maps remain fairly uniform, suggesting limited specialization in how they combine their inputs.
This may help explain their similar performance to the MLP and suggests that realizing the benefits of more
expressive connectors may require further training, changes to their optimization, or new residue-specific training tasks.

\begingroup
\setlength{\abovecaptionskip}{3pt}
\resultfig{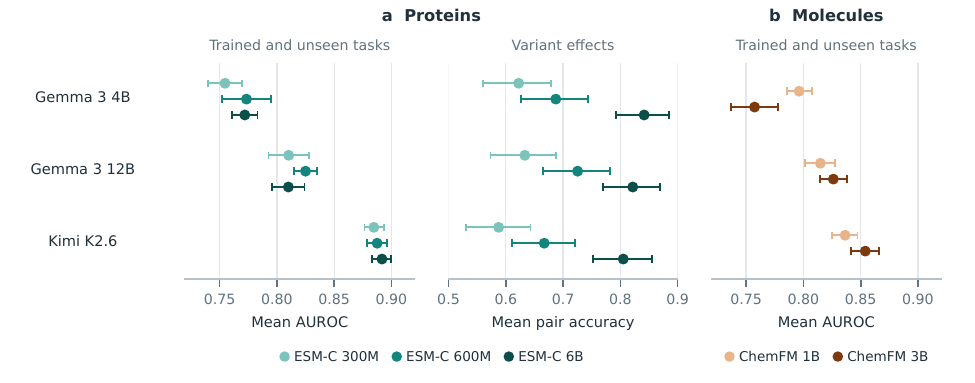}{\textbf{Specialist and LLM scaling.}
\textbf{(a)}~Proteins: mean task AUROC and variant pair accuracy; \textbf{(b)}~molecules: mean task AUROC.
Protein property performance varies more across LLMs than across specialist sizes, while variant effects
show the opposite pattern. Error bars show 95\% confidence intervals, including seed variation for Gemma 3 4B and 12B.}
\endgroup

\paragraph{Specialist and LLM scaling.}
We next examine how performance depends on the scale of both the specialist and the LLM, considering
proteins (Figure~\ref{fig:results_scaling}a) and molecules (Figure~\ref{fig:results_scaling}b).
For proteins, we compare ESM-C models from 300M to 6B parameters across the three LLMs. Protein property
performance varies more across LLMs than across specialist sizes. This may partly reflect the larger difference
in scale between the LLMs: Kimi K2.6 has 1T total and 32B active parameters, compared with 4B or 12B for
Gemma 3. Architecture and pretraining also differ, so this comparison does not isolate parameter count.

To test how scaling affects sensitivity to small sequence changes, we also evaluate variant effects.
We compare pairs of mutations from the same protein with opposite pathogenicity or loss of function labels.
Pair accuracy measures how often the positively labeled variant receives the higher Yes/No log odds,
with ties contributing one half.

This evaluation reveals a different scaling pattern. Increasing ESM-C from 600M to 6B yields substantial gains across
all three LLMs, whereas increasing LLM size does not consistently improve performance. The relative benefit
of specialist and LLM scaling therefore depends on the task: a more capable LLM can make better use of
available specialist information, but the specialist must first encode the distinctions needed to answer
the question.

\begingroup
\setlength{\abovecaptionskip}{3pt}
\resultfig{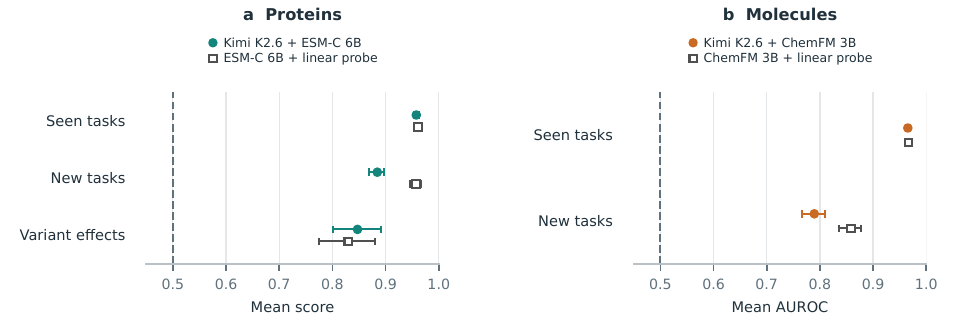}{\textbf{Comparison with supervised linear probes.}
Kimi K2.6 results using the four-token MLP connector of the previous section.
On trained tasks, connectors match logistic regression in both domains.
On new tasks, connectors trail the supervised probes, but are evaluated zero-shot.
Scores are mean task AUROC or variant pair accuracy; error bars show 95\% bootstrap intervals.}
\endgroup

\paragraph{Comparison with linear baselines.}
The preceding results establish that LIP supports nontrivial prediction across diverse tasks. To put this
performance in context, we compare the LLM connectors with supervised linear probes: logistic regression
classifiers trained to predict each task's labels directly from the same frozen specialist embeddings, without an LLM.
On trained tasks, the LLM connector matches the supervised probe in both domains
(Figure~\ref{fig:results_baseline_comparison}).

On new tasks, the LLM connectors underperform the probes in both domains, but this is not a comparison
under matched supervision. Each probe is trained on labels for the task being evaluated, whereas the connector
must answer these questions without explicit training on those tasks. The probes therefore provide a
supervised reference for the information accessible in the specialist embeddings, while the connectors are
being tested on zero-shot task transfer.
For protein tasks, mean zero-shot AUROC is 0.88, compared with 0.96 for the supervised probe.

\FloatBarrier

%% file: sections/discussion.tex
\section{Discussion and outlook}\label{sec:discussion}

LIP demonstrates a direct, high-bandwidth channel from scientific specialists to a shared language model: one latent
input supports many questions, and independently aligned inputs can be used together. The broader position is that
communication between models should not require every useful property to be selected and rendered for a human
reader first. Keeping the LLM and its existing connections fixed makes extension possible through independent connector
training, allowing a new specialist to join without changing those already connected.

The demonstrations establish three forms of reuse. Connections retain useful information about
objects excluded from connector fitting; fixed connections support questions beyond their training tasks; and separately
trained connectors support joint questions in combinations never fitted together, including protein--compound
relationships. The quality of these predictions depends on both LLM capability and specialist
scale. More capable LLMs generally provide better backbones for trained and zero-shot tasks, with the clearest
gains in task transfer and composition. On trained tasks, the four token MLP connectors match supervised
linear probes in both domains.

\paragraph{Limitations.}\label{sec:limitations}
Despite these gains, our experiments explore only a small part of the possible
connector designs. More expressive transformer connectors do not yet outperform MLPs, and their attention
remains fairly uniform after 2,000 training steps (Section~\ref{sec:results-ablations-expansion}). Further
extensions must account for the extra inference cost of additional tokens and, for residue-level or spatial
questions, the need to preserve finer detail in specialist representations.

Our scaling comparisons span LLMs with different model families and training histories,
so they do not isolate parameter count as the cause of improved performance. Experiments spanning more specialists, scientific domains and LLMs will also be needed to establish
the breadth of the framework's usefulness. Evaluating these connections in scientific workflows would help
determine how their benefits compare with the costs of constructing and maintaining them.

Our connectors also supply specialist representations to an LLM that was not
trained to receive them. Connector training can improve performance on supervised tasks while reducing transfer, as observed for
molecules. Together with the LLM's uneven use of specialist information during reasoning and composition, this
points to a remaining challenge: enabling independently connected specialists to contribute
reliably to the LLM's answers across changing questions and combinations.

\paragraph{Outlook.}\label{sec:outlook}
Preparing LLMs for representation exchange during their own training could enable more effective use of specialist information. Beyond
learning to receive latent inputs, models could learn to reason in latent space, working with complex
concepts that are difficult to express in human language. Emerging approaches to continuous and
modality-grounded reasoning provide starting points for exploring this
possibility~\cite{hao2024coconut,li2025latentvisual}. Such capabilities would extend the benefit of direct
communication into the reasoning that follows it.

For LLM developers, concrete steps include pretrained slots for external representations, extending the
reserved placeholders that already carry numerical values~\cite{golkar2024xval} or a multilingual encoder's
sentence embedding~\cite{agarwal2025languages}, and training
objectives that reward their use across questions and combinations of sources. Preparing a shared LLM
in this way could make it more receptive to specialists connected later through independent alignment,
preserving extensibility without retraining the LLM for each addition. Documented, versioned injection
interfaces would help sustain these connections as models evolve, making connectors easier to validate
and refit across releases.

Specialist developers can contribute consistent extraction interfaces and clear descriptions of what their
representations encode, including their granularity. These conventions would make models easier to connect
while leaving them free to learn different representations. Connectors would supply the alignment,
preserving the diversity of information available to the LLM.

Together, these efforts could support a growing registry of independently connected specialists alongside
existing tool interfaces (Figure~\ref{fig:hero}b). MCP, retrieval and Python would remain indispensable for
accessing sources, running computations and acting on the world; LIP would enrich the information available
within the agent's context. We call on the community to develop models and interfaces that let scientific
agents receive and reason with specialist knowledge at the bandwidth its representations support.

\section*{Acknowledgments}

We would like to acknowledge the support of the Simons Foundation and of
Schmidt Sciences. This work was supported in part by the AI2050 program at Schmidt Sciences (Grant G-25-70028). Additionally, computations were run at facilities supported by the Scientific Computing Core at the Flatiron
Institute, a division of the Simons Foundation. We would like to thank Fran\c{c}ois Lanusse, Irina Espejo Morales, Henri Palacci, and Roshan Rao for inspiring conversations pertaining to this project.

%% file: sections/appendix.tex
\clearpage
\section*{Supplementary material}

\makeatletter
\setcounter{tocdepth}{2}
\@starttoc{stoc}
\let\supp@addcontentsline\addcontentsline
\renewcommand{\addcontentsline}[3]{%
  \def\supp@ext{#1}\def\supp@main{toc}%
  \ifx\supp@ext\supp@main\supp@addcontentsline{stoc}{#2}{#3}%
  \else\supp@addcontentsline{#1}{#2}{#3}\fi}
\makeatother
\bigskip

\input{sections/supplementary_methods}
\FloatBarrier
\input{sections/supplementary_evidence}
\FloatBarrier

%% file: sections/supplementary_methods.tex
\section{Methods: building a connector}\label{sec:supp-methods}
This section gives the construction recipe used for our demonstration. The choices of representation, alignment
reference, injection layers and training objective are implementation choices within LIP. The shared evaluation
protocol is in Section~\ref{sec:supp-protocol}; additional controls are in Section~\ref{sec:supp-evaluation}.

\subsection{Models and representations}\label{sec:supp-models}
We connect each specialist independently to Gemma 3 4B, Gemma 3 12B and Kimi K2.6. All pretrained weights are frozen,
and specialist embeddings are extracted before connector fitting. Table~\ref{tab:connector_instances} specifies the
selected layers, pooling and concatenation order; the same specialist representation is used for all three LLMs.

For proteins, we use MIMIC, ESM-C 600M and PGLM 1B. The ESM-C and PGLM revisions are
\texttt{fcc9d36f} and \texttt{676c1bcb}, respectively. Mean pooling excludes special and padding tokens. MIMIC accepts sequences
up to 10,000 amino acids. Longer inputs to ESM-C and PGLM are split into windows of at most 2046 and 1023 amino
acids, respectively, and window means are weighted by their lengths.

The molecule models are MoLFormer-c3 (\texttt{9f1b9ea3}), ChemFM-1B (\texttt{f99dc2e8}) and Uni-Mol2 1.1B
through \texttt{unimol\_tools} 0.1.6. MoLFormer and ChemFM use SMILES, derived from InChI when necessary, truncated
to their configured limits of 202 and 512 tokens. Uni-Mol2 uses coordinates with hydrogens removed. Its current
extraction retries failed three-dimensional conformer generation from random coordinates, checks specified
stereochemistry, and otherwise retains the original fallback. The current Uni-Mol2 extraction uses isolated padding and no atom cropping. These extraction choices are part of the connector's representation definition.

\subsection{Data, splits and supervised targets}\label{sec:supp-data}
\subsubsection{Sources and object separation}\label{sec:supp-splits}
Protein sequences and annotations come from reviewed UniProtKB/Swiss-Prot entries, release 2026\_03, and compounds
from three-star ChEBI entries, release 255. Text targets use the corresponding annotation fields with source
citations and markup removed. Definitions, classes, roles and functional-parent relations are processed separately
to control which information is used for fitting. Rhea supplies reaction relationships for cross-domain evaluation.

Protein splits use MMseqs2 at 30\% sequence identity and 80\% coverage of both sequences, checked in both search
directions at $E\leq10^{-3}$. Related chemical forms are grouped before molecule splitting, with similarity links
at ECFP4 Tanimoto similarity 0.5 (radius 2, 2048 bits). Cross-split links are removed by reassignment or exclusion;
only objects with the required embeddings and annotations enter fitting. These exclusions apply to every object
in a training prompt, including worked examples and relational partners. They concern connector training;
overlap with LLM or specialist pretraining is unknown.

\subsubsection{Supervised and reserved targets}
Table~\ref{tab:supp-targets} summarizes the task partition. Names provide alignment references and textual
controls, but naming the object is not a training target. Annotation-derived negatives indicate missing
annotations, not verified biological or chemical absence. A question withheld from task training may still be
answerable from descriptions or correlated annotations (Section~\ref{sec:supp-zero-shot-exposure}).

\begin{table}[tbp]
\centering\small
\caption{Task supervision and reserved categories. Reserved questions may have related information in description targets.}\label{tab:supp-targets}
\begin{tabularx}{\linewidth}{@{}l Y Y@{}}
\toprule
Domain & Task supervision & Reserved question categories \\
\midrule
Protein & Function, family, subunit, catalytic activity, EC class, pathway, location and selected keywords;
relations and variant effects, pathogenicity and loss of function. & Lineage, length, selected modifications,
cofactors and domains. \\
Compound & Definitions, mechanism roles, structural classes and functional parents; relations between compounds.
& Selected provenance, application and chemical roles; formula, mass, charge and element-based questions. \\
\bottomrule
\end{tabularx}
\end{table}

Molecule roles are separated by ontology ancestry before the remaining mechanism roles are assigned to training;
class labels include transitive \texttt{is\_a} ancestry. The partition accounts for related ontology labels as well as the named categories themselves. The mass threshold is fixed at 329.29 Da, the median of the original training split, and retained after subsequent data filtering.

\subsubsection{Sampling and relational examples}\label{sec:supp-sampling}
Protein training draws 40\% single-object text, 20\% Yes/No, 20\% relational and 20\% variant examples. Molecule
training draws 50\% text, 25\% Yes/No and 25\% relational examples (Figure~\ref{fig:data_figure}). For ordinary
object sampling, an object's weight is proportional to $1/\sqrt{n_c}$, where $n_c$ is its cluster size. Yes/No
sampling balances positive and negative labels after selecting a question type. The variant stream includes
mutation-effect descriptions, pathogenicity and loss-of-function questions, and wild-type--mutant comparisons.

Relational answers state annotation-derived facts about one, two or three objects and identify them by prompt
position. Half of odd-one-out draws use hard negatives differing at the annotation level requested. Every injected
slot within a training example uses the same connector; combining independently trained connectors is reserved
for evaluation.

\subsection{Question examples and prompt construction}\label{sec:supp-prompts}
The examples below illustrate the question forms used in our experiments. \texttt{<SLOT>} denotes an injected position. Chat-template tokens and worked-example preambles are
omitted; the displayed targets illustrate annotation supervision, not generated responses. Descriptions and
binary questions concern one object, relational questions compare objects, and combined representations and
cross-domain questions test composition of independently trained connectors.

\begin{quote}\small\raggedright
\textbf{Protein description.}\\*
\texttt{What does UniProt say about its subunit structure? The protein is<SLOT>.}\\*
Target: The complex is composed of two ATP-binding proteins, two transmembrane proteins and a solute-binding protein.

\smallskip
\textbf{Compound description.}\\*
\texttt{What is the compound given as<SLOT> active as?}\\
Target: Role: antifungal agent.

\smallskip
\textbf{Yes/No property question.}\\*
\texttt{Here is an unknown protein, given to you as an embedding:<SLOT>. Does this protein bind DNA? Answer Yes or No.}\\*
For compounds, the same frame asks, for example, \texttt{Is this compound a lipid?}, \texttt{Is this compound a
metabolite?}, or \texttt{Is the molecular mass of this compound above 329 Da?}

\smallskip
\textbf{Several properties in one question.}\\*
\texttt{Give a short profile of the protein<SLOT>: its family, where it is found and what it does.}

\smallskip
\textbf{Relational comparison.}\\*
\texttt{Do the proteins<SLOT> and<SLOT> belong to the same family? Explain.}\\*
A compound example is \texttt{Do<SLOT> and<SLOT> derive from the same parent compound?}

\smallskip
\textbf{Relational choice.}\\*
\texttt{Three proteins:<SLOT>,<SLOT>,<SLOT>. Two share a family. Which one does not?}\\
Target: The first and third belong to the CDC42SE/SPEC family; the second belongs to the CTF8 family. The second is
the odd one out.

\smallskip
\textbf{Variant effect.}\\*
\texttt{Here is a mutant protein, given to you as an embedding:<SLOT>. What is the effect of its mutation?}

\smallskip
\textbf{Variant pathogenicity and loss of function.}\\*
\texttt{Is the following missense variant pathogenic? Variant:<SLOT>}\\
\texttt{Consider the protein variant<SLOT>. Does this variant lose activity?}\\
Both use a binary answer convention.

\smallskip
\textbf{Wild-type--mutant comparison.}\\*
\texttt{Protein 1:<SLOT>. Protein 2:<SLOT>. Protein 2 is a point mutant of protein 1. What is the effect of the mutation?}

\smallskip
\textbf{Several representations of one object.}\\*
\texttt{Here is an unknown protein, given to you as an embedding:<SLOT><SLOT><SLOT>. Does this protein bind DNA?
Answer Yes or No.}\\*
The slots contain different specialists' representations of the same protein.

\smallskip
\textbf{Cross-domain question.}\\*
\texttt{Here is an unknown protein, given to you as an embedding:<SLOT>. Here is an unknown compound, given to
you as an embedding:<SLOT>. Does this protein catalyze a reaction in which this compound is a substrate or a
product? Answer Yes or No.}

\smallskip
\textbf{Name elicitation.}\\*
\texttt{Here is an unknown protein, given to you as an embedding:<SLOT>. What is the name of this protein?
Reply with the protein name only.}\\*
The response can replace the embedding in a subsequent property question; a supplied-name control instead uses
the database name.
\end{quote}

Text and Yes/No training prompts vary wording, slot placement and requested answer format through a compositional
grammar. Worked examples use other training objects outside the query's cluster, and every injected example uses
the same connector. Selected phrasings are reserved for evaluation. Text and variant targets are capped at 48
tokens and relational targets at 96; truncation does not add an end-of-answer token. Binary evaluations use a
single user turn and the Yes/No frame above, with reasoning disabled unless explicitly tested. Requests for
reasoning change the response procedure described in Section~\ref{sec:results-reasoning}.

\subsection{Align and Select}\label{sec:supp-construction}

\subsubsection{Specialist read profiles}\label{sec:supp-read-probes}
Before alignment, ridge probes compare specialist layers using property labels and five cluster-separated folds.
For proteins, additional clinical and functional variant probes use the difference between mutant and wild-type
representations. These probes assess the specialist representation without using the LLM. Candidate layers and
concatenations are compared by task performance, leading to the human-selected representations in
Table~\ref{tab:connector_instances} and those used in the scaling experiments.

Figures~\ref{fig:supp_read_profiles_protein} and~\ref{fig:supp_read_profiles_molecule} show layer profiles
alongside the chosen representations. Property means use the per-specialist task subsets retained in
the original scan summaries, applying the same subset to every representation of that specialist. The subsets
differ across specialists, so these curves primarily compare layers within a specialist. Protein variant curves
average the clinical and functional probe AUROCs. ChemFM 3B uses all 59 property probes and concatenates
the last-token and mean-pooled embeddings at each layer. The profiles are exploratory guidance for representation choice.

\resultfig{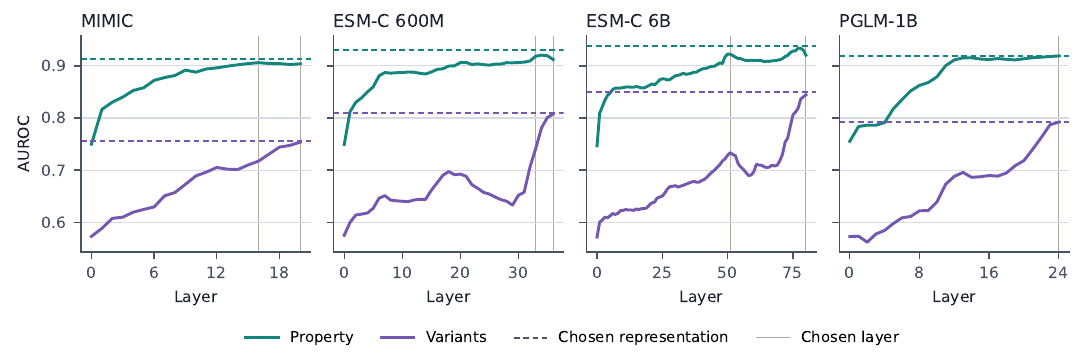}{\textbf{Protein specialist read profiles.}
Mean AUROC of property (green) and variant (purple) probes for mean-pooled representations across specialist layers.
Vertical lines mark chosen layers, and dashed lines show the corresponding scores of the chosen single-layer or
concatenated representation.}

\resultfig{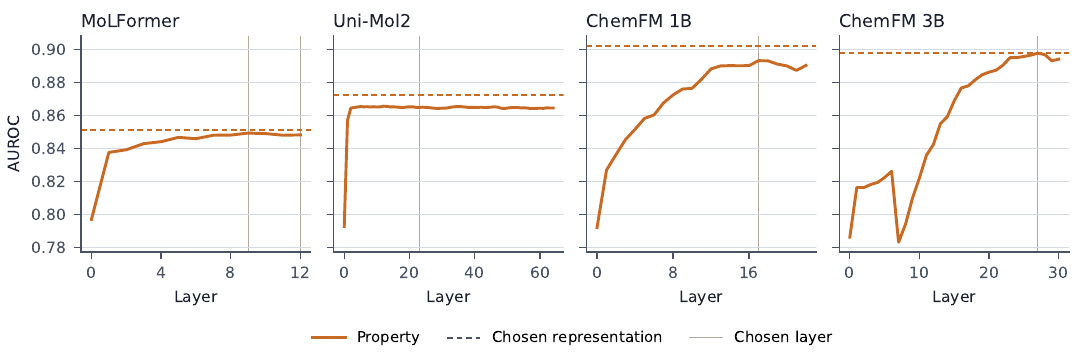}{\textbf{Molecule specialist read profiles.}
Mean property-probe AUROC across specialist layers, using mean-pooled embeddings for MoLFormer, Uni-Mol2 and
ChemFM 1B, and concatenated last-token and mean-pooled embeddings for ChemFM 3B.
Vertical lines mark chosen layers, and dashed lines show the scores of the full chosen representations.}

\FloatBarrier
\subsubsection{Standardization and the write operation}\label{sec:supp-write-operation}
Coordinate means and standard deviations are fitted on the regression objects and fixed thereafter:
$z_j=(e_j-\mu_j)/(s_j+10^{-8})$. At selected layer $\ell$, an affine map
$\delta_\ell=W_\ell z+b_\ell$ supplies the slot's write direction. Normalization and mixing
(Equation~\ref{eq:mix}) use float32 with norm denominators bounded below by $10^{-6}$ before casting back to the
LLM's computation dtype. A learned object-independent constant supplies the input-embedding slot direction.
Other token positions undergo ordinary LLM computation.

\subsubsection{Reference states and ridge fitting}\label{sec:supp-ridge-fitting}
Reference states are captured in prompts containing an object's description followed by its name. Protein targets
average the name token states; molecule targets use the final name token. For Kimi, the target for write layer
$\ell>0$ comes from layer $\max(\ell,p_*)$, with $p_*=13$ for proteins and 9 for molecules; these reference depths
were selected using the same diagnostic AUROCs. Gemma targets use the matching layer. This reference-depth rule
sets the regression target and does not restrict where the connector may write.

Ridge maps fit centered reference states from standardized specialist representations, with the intercept set
by the reference mean. The penalty is selected by five-fold cross-validation over 28 logarithmically spaced
values from $10^0$ to $10^9$. The scan objects are excluded from both fitting and standardization for selection
maps; the production maps use the full fitting set. Reference capture uses names only during construction,
not when the connector receives a new object.

\subsubsection{Name states as an alignment reference}\label{sec:supp-studies}\label{sec:supp-reference-states}
We compared final name-token states with mean description states from either bare
text or an assistant response following a prompt containing the name (Figure~\ref{fig:supp_alignment_targets}).
Ridge maps used the same 9017 fitting proteins; injected-task performance was averaged over five Yes/No questions
on 200 held-out proteins. Name-state predictions gave better injected-question performance. This illustrates why alignment quality is assessed through
the LLM's responses.

\resultfig{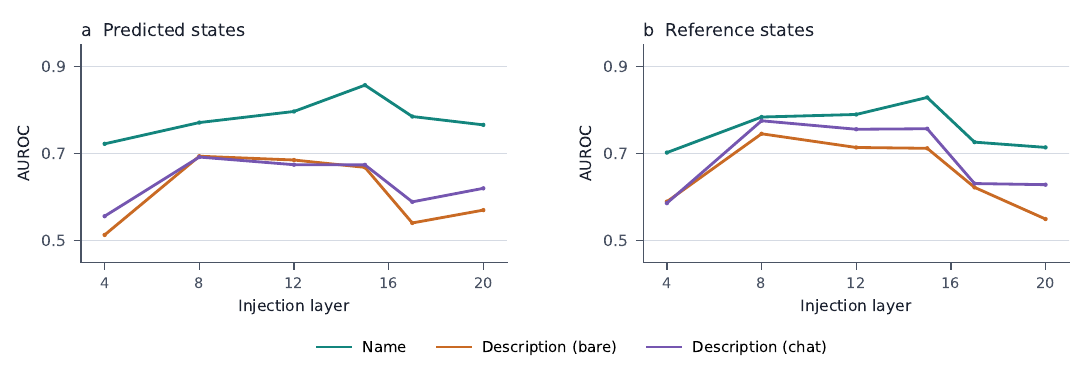}{\textbf{Name and description states as alignment targets.}
Mean task AUROC at each scanned layer in the informative band of Kimi K2.6, injecting
\textbf{(a)}~ridge-predicted states or \textbf{(b)}~the LLM's actual reference states.}

\subsubsection{LLM write profiles and selection}\label{sec:supp-write-selection}
Each ridge prediction is injected separately at each LLM layer. Mean AUROC across five diagnostic questions gives
a curve for each connector, alongside a curve obtained by injecting the LLM's own reference states
(Figure~\ref{fig:supp_write_profiles}). The questions concern RNA binding, transcription regulation, hydrolase
activity, membrane association and Pseudomonadota for proteins, and benzenoid aromatic compounds, esters, lipids,
drug roles and net charge for molecules. These questions guide layer selection.

For each domain and LLM, local maxima of the maximum envelope over connector and reference curves guide the
choice of layers spanning depth (Table~\ref{tab:llm_writes}). The shared schedule is a choice for this demonstration.
The figure shows the three specialists in each domain from Table~\ref{tab:connector_instances} and the maximum
across those curves and the own-state reference. The original Kimi protein selection also considered an ESM-C 6B
curve, which is omitted from this display; the selected layers are unchanged.

\resultfig{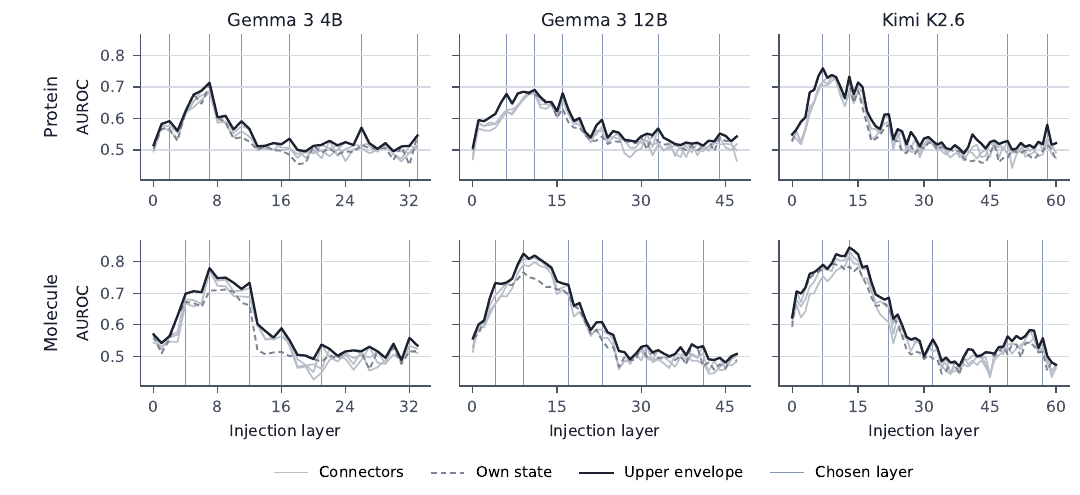}{\textbf{LLM injection-layer profiles.}
Mean diagnostic-task AUROC from single-layer injection of aligned connector outputs (gray) and the LLM's own
reference states (dashed), with the maximum across the displayed curves in black and chosen layers marked vertically.
Rows show proteins and molecules; columns show Gemma 3 4B, Gemma 3 12B and Kimi K2.6.}

Mixing weights are initialized from each connector's above-chance mean AUROC on the trained diagnostic questions,
normalized by its best layer and clipped to $[0.1,0.9]$, then learned during training.

\FloatBarrier
\subsection{Train}\label{sec:supp-training}
\subsubsection{Objective and token weights}
Only the connector's affine maps, biases, input constant and mixing weights are trained. Gradients pass through
the frozen LLM from answer tokens to these parameters. With teacher-forced negative log likelihood $\ell_{it}$
and answer-token weight $w_{it}$, the objective is
\begin{equation}
\mathcal J=\frac{\sum_{i,t}w_{it}\ell_{it}}{\sum_{i,t}w_{it}}.
\end{equation}
Prompt tokens receive no loss. Position-dependent weights emphasize annotation content whose prediction benefits
from knowing the object's name; connecting text and end tokens receive a floor weight. Each factual span of a
relational answer restarts its annotation-specific position weights, and binary decision tokens receive full
weight before stream-specific scaling.

\subsubsection{Optimization and batching}
We use AdamW with a separate base learning rate for each map-weight tensor, bias vector and input constant,
set to 0.002 times the root mean square of that tensor's initial values. Across the 18 connectors, the resulting
absolute learning rates range approximately from $3\times10^{-7}$ to $2\times10^{-2}$ for map weights,
$8\times10^{-5}$ to 3 for biases, and $4\times10^{-6}$ to $6\times10^{-4}$ for input constants;
mixing logits use a learning rate of 0.01. A 20-step linear warmup precedes a constant learning rate;
gradients are clipped to global norm 1.0. Per-step relative weight decay of $2.5\times10^{-4}$ applies to the maps, scaled by warmup and independent
of learning rate; biases, the input constant and mixing logits are excluded. Each optimizer update targets 85 examples, with microbatches where needed to fit
memory. Main comparisons use the final checkpoint after 2,000 updates, without selecting an intermediate
checkpoint by evaluation performance. Main experiments use one training seed per connector; the scaling comparison includes the repeated runs described in Supplementary Section~\ref{sec:supp-protocol}.

\subsubsection{Training through the frozen LLM}
Kimi training adapts the expert-parallel implementation described by Workshop Labs for
Trellis~\cite{foote2026trellis}, retaining INT4 expert weights with bfloat16 computation; Gemma uses a
single-GPU bfloat16 backend. Backpropagation computes gradients
through the LLM with respect to its inputs, without LLM weight gradients or optimizer states. Packed examples
retain separate causal-attention boundaries and restart position indices for each example. Section~\ref{sec:setup-cost}
gives representative training costs.

\subsection{Applying and combining connectors}\label{sec:supp-interface}
Each slot selects a connector and an embedding extracted according to its representation definition. The slot
retains its own normalization, maps, write layers and mixing weights; where that connector has no write, the
LLM proceeds normally. Different connectors need not share injection schedules. Several slots may encode one
object through different specialists or supply different objects, without combining or refitting connector parameters.

A checkpoint identifies its LLM, specialist, extraction procedure, normalization and write schedule. These must
match at inference because the maps are specific to the model pair and representation. All inference backends
apply the same recorded writes at the specified slots and layers.

%% file: sections/supplementary_evidence.tex
\section{Evaluation methods and supporting evidence}\label{sec:supp-evaluation}
The following methods and controls address whether the LLM uses the supplied object, transfers beyond supervised
questions, combines independently trained connectors and accesses information beyond names.

\input{sections/supplementary_protocol}

\subsection{Within-task generalization}\label{sec:supp-individual}
\paragraph{Unseen question wording.}\label{sec:supp-wording}
Selected question phrasings are withheld during connector training. A matched control asked the same
questions about the same proteins using either training or reserved wording. Paired confidence intervals for changes in the
information-access gap included zero for each protein connector, with no detected loss under the reserved
wording. This controls wording while holding the object, target and context fixed.

\paragraph{Task groups.}
Displayed groups use fixed task memberships and equal-weight means of task AUROCs. Protein groups collect related
properties; molecule groups use chemical meaning and ChEBI ancestry. These are summaries of correlated questions,
not independent dimensions of information. Rare-label limitations follow Section~\ref{sec:supp-protocol}.

\subsection{Transfer to new questions}\label{sec:supp-transfer}
\subsubsection{Description exposure}\label{sec:supp-zero-shot-exposure}
Withholding a task question does not remove its answer from all other supervision. We estimate how often a
positive object's description target conveys the queried property, using the training sampler's object and
text-stream weights and counting repeated presentations relative to all training examples. Property cues are
matched in capped descriptions: a 200-character approximation for proteins and 48 Kimi tokens for molecules.

Low-exposure question types have estimated exposure below 0.1\%; the main figures also identify starred types
with exposure up to 0.32\%. These estimates concern description targets, not all routes by which information
could enter supervision: variant-effect text, relational answers and worked context are outside this metric,
and correlated or implicit clues may remain. Read-layer probes and write-selection diagnostics can also expose
labels for questions withheld from connector task training. Thus task transfer does not imply that the complete
construction procedure was blind to every tested property.

\subsubsection{Tasks and populations}\label{sec:supp-transfer-populations}
Transfer questions concern protein lineage and properties, compound roles and mass. Molecule transfer uses held-out structural groups without a training analog at ECFP4 Tanimoto
similarity 0.5. The mass threshold is fixed at 329.29 Da, the median of the original training split, and retained after subsequent data filtering. Protein length uses 320 amino acids.
Group means weight questions equally, with lineage labels nested and correlated. Each comparison is restricted
to objects eligible for its tasks and input conditions.

\subsection{Effect of training}\label{sec:supp-transfer-limits}
Figure~\ref{fig:results_training_comparison} compares the same connector after \emph{Align} and \emph{Select} and
after \emph{Train}, on matched questions and objects. This isolates the change during connector task training
while holding the specialist and LLM fixed. It does not isolate individual supervision streams or distinguish which correlated training targets caused a transfer gain or loss.

\subsection{Merging representations of one protein}\label{sec:supp-same-object}
The composition comparison supplies the same protein through adjacent slots from MIMIC, ESM-C and PGLM, without
joint fitting. Repeated-source triples (MMM, EEE and PPP) control for the effect of supplying more slots without
adding a different specialist. Figure~\ref{fig:supp_composition_controls} compares the three different specialists
with the best single connector and the best repeated-source triple. Each comparator is selected separately for
trained and unseen tasks by mean task AUROC and held fixed across paired resamples.

\resultfig{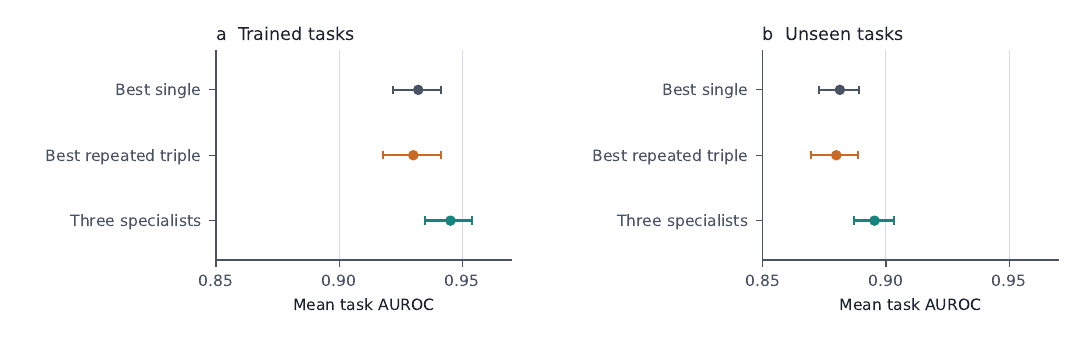}{\textbf{Complementary specialists improve information merging.}
Mean task AUROC with Kimi K2.6 on 2,036 proteins for \textbf{(a)}~five trained and \textbf{(b)}~eight unseen questions,
comparing three different specialists with the best single connector and the best triple repeating one specialist.
Intervals are 95\% paired protein-bootstrap intervals (2,000 draws).}

These composition comparisons are exploratory. A separate prespecified test required improvement over both controls on
at least two of three metrics, with paired intervals above zero; that criterion was not met.

\subsection{Protein--compound relations}\label{sec:supp-cross-domain}
\paragraph{Tasks and populations.}
Rhea reaction participation asks whether a protein catalyzes a reaction in which the compound is a substrate or
product. BindingDB supplies protein--compound binding questions. Relations stated in training descriptions are
excluded, and populations distinguish whether each object appeared in its own connector's training pool.
The Rhea comparison in Figure~\ref{fig:results_merging}c uses unseen proteins with familiar or unseen compounds.
The connectors are fitted independently, without cross-domain question supervision. Annotation-negative pairs
indicate absent annotations rather than verified biochemical non-participation. These population comparisons are exploratory; differences between them
do not isolate the causal effect of training exposure.

\paragraph{Matched groups and ordering.}
Each group starts from two annotated protein--compound pairs; swapping compounds yields two mismatched pairs made
from the same objects. Each annotated pair is compared with each mismatched pair sharing its protein or compound,
using the Yes/No log-probability contrast. Correct orderings score one, ties one half and incorrect orderings zero;
the four comparisons are averaged within a group, then across groups. A score additive in separate protein and
compound contributions averages exactly 0.5, so above-chance ordering requires a relationship between inputs.

\subsection{Information beyond names}\label{sec:supp-beyond-name}
\paragraph{Replacing embeddings with elicited names.}
The LLM first generates a name from the protein embedding. That response then replaces the slot in the same
property questions. Every eligible object remains, whether or not its generated name is correct. Comparisons use
matched objects and question sets and average task AUROCs equally. A decline after replacement can reflect an
incorrect name as well as information lost by compressing the embedding to a name.

\paragraph{Identical-name groups.}
Objects are grouped by the exact supplied database-name string. Within-name concordance compares positive and
negative objects only within such groups, counting ties as one half and pooling pairs across groups for each question.
Question-level concordances are then averaged equally. A deterministic supplied-name score is
identical for a fixed question within each group and therefore scores 0.5. Above-chance concordance establishes
that the embedding provides distinctions unavailable from that shared name alone. Figure~\ref{fig:results_reasoning}b--c
applies this test to four low-exposure lineage questions and to protein length above 320 amino acids.
Only groups containing both labels are informative; nested lineage labels and limited informative groups constrain
the interpretation. Uncertainty resamples name groups rather than treating all within-group pairs as independent.

\FloatBarrier

%% file: sections/supplementary_protocol.tex
\subsection{Evaluation protocol}\label{sec:supp-protocol}
\paragraph{Objects and tasks.}
Connector training and evaluation use the object exclusions described in Section~\ref{sec:supp-data}.
Read and write selection use training-pool objects; write-scan objects are excluded from fitting the selection
maps but are not a held-out evaluation population. Monitoring and final evaluations are distinguished from these
construction scans. Cross-domain populations additionally distinguish familiar and unseen objects in each domain. Matched comparisons use
the same eligible objects and questions for the conditions being compared. Enrichment for uncommon labels changes
the evaluated population relative to a random sample.

\paragraph{Input conditions.}
The \emph{own} condition supplies the queried object's embedding. A \emph{shuffled} control substitutes a different
object's embedding while retaining the question, target and connector. The \emph{placeholder} condition leaves
the slot uninjected. The \emph{name} condition replaces it with the database name as text; names can activate LLM prior knowledge but are not an upper bound on specialist information. An \emph{initial} connector has completed
\emph{Align} and \emph{Select}, before answer-loss training. Question examples appear in Section~\ref{sec:supp-prompts}.

\paragraph{Scores.}
For binary questions, the score is $s_i=\log p(\mathrm{Yes}\mid x_i)-\log p(\mathrm{No}\mid x_i)$ at the answer
position, with AUROC computed against task labels. Group means weight their tasks equally; chance is 0.5.
For a supplied reference answer, the information-access gap is
\begin{equation}\label{eq:supp-gap}
\Delta_i=L_i^{\mathrm{shuf}}-L_i^{\mathrm{own}},
\end{equation}
where $L_i$ is mean answer-token negative log likelihood. A positive gap means that the correct embedding helps
predict that reference; it does not measure the accuracy of generated text. Relational ordering and within-name
concordance are defined with their controls in Section~\ref{sec:supp-evaluation}.

\paragraph{Uncertainty.}
Where intervals are shown, matched conditions share resampled objects, and task averages share the same resample
across tasks. Within-name comparisons resample name groups. In Figure~\ref{fig:results_scaling}, Gemma protein property and ChemFM 3B molecule results average three training seeds; their approximate 95\% intervals combine object sampling variance with the between-seed variance of the mean.
The remaining conditions in that figure use one seed and percentile bootstrap intervals. Variant intervals resample parent protein clusters. Rare labels can leave some resamples without both classes and therefore without a defined AUROC;
narrow intervals in such cases do not establish precise performance.

%% file: refs.bib
@misc{perozzi2024graph,
  title = {{Let Your Graph Do the Talking: Encoding Structured Data for LLMs}},
  author = {Bryan Perozzi and Bahare Fatemi and Dustin Zelle and Anton Tsitsulin and Mehran Kazemi and Rami Al-Rfou and Jonathan Halcrow},
  year = {2024},
  eprint = {2402.05862},
  archivePrefix = {arXiv},
  url = {https://arxiv.org/abs/2402.05862}
}

@misc{girdhar2023imagebind,
  title = {{ImageBind: One Embedding Space To Bind Them All}},
  author = {Rohit Girdhar and Alaaeldin El-Nouby and Zhuang Liu and Mannat Singh and Kalyan Vasudev Alwala and Armand Joulin and Ishan Misra},
  year = {2023},
  eprint = {2305.05665},
  archivePrefix = {arXiv},
  url = {https://arxiv.org/abs/2305.05665},
  note = {CVPR 2023}
}

@misc{panagopoulou2023x,
  title = {{X-InstructBLIP: A Framework for aligning X-Modal instruction-aware representations to LLMs and Emergent Cross-modal Reasoning}},
  author = {Artemis Panagopoulou and Le Xue and Ning Yu and Junnan Li and Dongxu Li and Shafiq Joty and Ran Xu and Silvio Savarese and Caiming Xiong and Juan Carlos Niebles},
  year = {2023},
  eprint = {2311.18799},
  archivePrefix = {arXiv},
  url = {https://arxiv.org/abs/2311.18799},
  note = {ECCV 2024}
}

@misc{carrami2024pqa,
  title = {{PQA: Zero-shot Protein Question Answering for Free-form Scientific Enquiry with Large Language Models}},
  author = {Eli M Carrami and Sahand Sharifzadeh},
  year = {2024},
  eprint = {2402.13653},
  archivePrefix = {arXiv},
  url = {https://arxiv.org/abs/2402.13653}
}

@misc{pham2023models,
  title = {{Let Models Speak Ciphers: Multiagent Debate through Embeddings}},
  author = {Chau Pham and Boyi Liu and Yingxiang Yang and Zhengyu Chen and Tianyi Liu and Jianbo Yuan and Bryan A. Plummer and Zhaoran Wang and Hongxia Yang},
  year = {2023},
  eprint = {2310.06272},
  archivePrefix = {arXiv},
  url = {https://arxiv.org/abs/2310.06272},
  note = {ICLR 2024}
}

@misc{liu2024prott3,
  title = {{ProtT3: Protein-to-Text Generation for Text-based Protein Understanding}},
  author = {Zhiyuan Liu and An Zhang and Hao Fei and Enzhi Zhang and Xiang Wang and Kenji Kawaguchi and Tat-Seng Chua},
  year = {2024},
  eprint = {2405.12564},
  archivePrefix = {arXiv},
  url = {https://arxiv.org/abs/2405.12564},
  note = {ACL 2024}
}

@misc{wang2024protchatgpt,
  title = {{ProtChatGPT: Towards Understanding Proteins with Large Language Models}},
  author = {Chao Wang and Hehe Fan and Ruijie Quan and Yi Yang},
  year = {2024},
  eprint = {2402.09649},
  archivePrefix = {arXiv},
  url = {https://arxiv.org/abs/2402.09649}
}

@misc{ramesh2025communicating,
  title = {{Communicating Activations Between Language Model Agents}},
  author = {Vignav Ramesh and Kenneth Li},
  year = {2025},
  eprint = {2501.14082},
  archivePrefix = {arXiv},
  url = {https://arxiv.org/abs/2501.14082},
  note = {ICML 2025}
}

@misc{kamai2026talking,
  title = {{Talking with the Latents -- how to convert your LLM into an astronomer}},
  author = {Ilay Kamai and Marc-Huertas Company and Mike J. Smith and Hagai B. Perets},
  year = {2026},
  eprint = {2602.09670},
  archivePrefix = {arXiv},
  url = {https://arxiv.org/abs/2602.09670}
}

@misc{fu2025cache,
  title = {{Cache-to-Cache: Direct Semantic Communication Between Large Language Models}},
  author = {Tianyu Fu and Zihan Min and Hanling Zhang and Jichao Yan and Guohao Dai and Wanli Ouyang and Yu Wang},
  year = {2025},
  eprint = {2510.03215},
  archivePrefix = {arXiv},
  url = {https://arxiv.org/abs/2510.03215},
  note = {ICLR 2026}
}

@misc{zou2025latent,
  title = {{Latent Collaboration in Multi-Agent Systems}},
  author = {Jiaru Zou and Ruizhong Qiu and Gaotang Li and Xiyuan Yang and Katherine Tieu and Pan Lu and Ke Shen and Hanghang Tong and Yejin Choi and Jingrui He and James Zou and Mengdi Wang and Ling Yang},
  year = {2025},
  eprint = {2511.20639},
  archivePrefix = {arXiv},
  url = {https://arxiv.org/abs/2511.20639},
  note = {ICML 2026}
}

@misc{yao2022react,
  title = {{ReAct: Synergizing Reasoning and Acting in Language Models}},
  author = {Shunyu Yao and Jeffrey Zhao and Dian Yu and Nan Du and Izhak Shafran and Karthik Narasimhan and Yuan Cao},
  year = {2022},
  eprint = {2210.03629},
  archivePrefix = {arXiv},
  url = {https://arxiv.org/abs/2210.03629},
  note = {ICLR 2023}
}

@misc{zhao2023chatbridge,
  title = {{ChatBridge: Bridging Modalities with Large Language Model as a Language Catalyst}},
  author = {Zijia Zhao and Longteng Guo and Tongtian Yue and Sihan Chen and Shuai Shao and Xinxin Zhu and Zehuan Yuan and Jing Liu},
  year = {2023},
  eprint = {2305.16103},
  archivePrefix = {arXiv},
  url = {https://arxiv.org/abs/2305.16103}
}

@inproceedings{tenney2019bert,
  title = {{BERT} Rediscovers the Classical {NLP} Pipeline},
  author = {Ian Tenney and Dipanjan Das and Ellie Pavlick},
  booktitle = {Proceedings of the 57th Annual Meeting of the Association for Computational Linguistics},
  year = {2019},
  publisher = {Association for Computational Linguistics},
  pages = {4593--4601},
  doi = {10.18653/v1/P19-1452},
  url = {https://aclanthology.org/P19-1452/}
}

@misc{meng2024deepstack,
  title = {{DeepStack: Deeply Stacking Visual Tokens is Surprisingly Simple and Effective for LMMs}},
  author = {Lingchen Meng and Jianwei Yang and Rui Tian and Xiyang Dai and Zuxuan Wu and Jianfeng Gao and Yu-Gang Jiang},
  year = {2024},
  eprint = {2406.04334},
  archivePrefix = {arXiv},
  url = {https://arxiv.org/abs/2406.04334},
  note = {NeurIPS 2024}
}

@misc{liu2024evollama,
  title = {{EvoLlama: Enhancing LLMs' Understanding of Proteins via Multimodal Structure and Sequence Representations}},
  author = {Nuowei Liu and Changzhi Sun and Tao Ji and Junfeng Tian and Jianxin Tang and Yuanbin Wu and Man Lan},
  year = {2024},
  eprint = {2412.11618},
  archivePrefix = {arXiv},
  url = {https://arxiv.org/abs/2412.11618}
}

@misc{agarwal2025languages,
  title = {{Languages are Modalities: Cross-Lingual Alignment via Encoder Injection}},
  author = {Rajan Agarwal and Aarush Gupta},
  year = {2025},
  eprint = {2510.27254},
  archivePrefix = {arXiv},
  url = {https://arxiv.org/abs/2510.27254}
}

@misc{kim2025mol,
  title = {{Mol-LLaMA: Towards General Understanding of Molecules in Large Molecular Language Model}},
  author = {Dongki Kim and Wonbin Lee and Sung Ju Hwang},
  year = {2025},
  eprint = {2502.13449},
  archivePrefix = {arXiv},
  url = {https://arxiv.org/abs/2502.13449},
  note = {NeurIPS 2025}
}

@misc{chen2026scicore,
  title = {{SciCore-Mol: Augmenting Large Language Models with Pluggable Molecular Cognition Modules}},
  author = {Yuxuan Chen and Changwei Lv and Yunduo Xiao and Zhongjing Du and Daquan Zhou and Yukun Yan and Zheni Zeng and Zhiyuan Liu},
  year = {2026},
  eprint = {2605.22287},
  archivePrefix = {arXiv},
  url = {https://arxiv.org/abs/2605.22287}
}

@misc{alayrac2022flamingo,
  title = {{Flamingo: a Visual Language Model for Few-Shot Learning}},
  author = {Jean-Baptiste Alayrac and Jeff Donahue and Pauline Luc and Antoine Miech and Iain Barr and Yana Hasson and Karel Lenc and Arthur Mensch and Katie Millican and Malcolm Reynolds and Roman Ring and Eliza Rutherford and Serkan Cabi and Tengda Han and Zhitao Gong and Sina Samangooei and Marianne Monteiro and Jacob Menick and Sebastian Borgeaud and Andrew Brock and Aida Nematzadeh and Sahand Sharifzadeh and Mikolaj Binkowski and Ricardo Barreira and Oriol Vinyals and Andrew Zisserman and Karen Simonyan},
  year = {2022},
  eprint = {2204.14198},
  archivePrefix = {arXiv},
  url = {https://arxiv.org/abs/2204.14198},
  note = {NeurIPS 2022}
}

@misc{li2023blip,
  title = {{BLIP-2: Bootstrapping Language-Image Pre-training with Frozen Image Encoders and Large Language Models}},
  author = {Junnan Li and Dongxu Li and Silvio Savarese and Steven Hoi},
  year = {2023},
  eprint = {2301.12597},
  archivePrefix = {arXiv},
  url = {https://arxiv.org/abs/2301.12597}
}

@misc{schick2023toolformer,
  title = {{Toolformer: Language Models Can Teach Themselves to Use Tools}},
  author = {Timo Schick and Jane Dwivedi-Yu and Roberto Dessì and Roberta Raileanu and Maria Lomeli and Luke Zettlemoyer and Nicola Cancedda and Thomas Scialom},
  year = {2023},
  eprint = {2302.04761},
  archivePrefix = {arXiv},
  url = {https://arxiv.org/abs/2302.04761}
}

@misc{kar2024brave,
  title = {{BRAVE: Broadening the visual encoding of vision-language models}},
  author = {Oğuzhan Fatih Kar and Alessio Tonioni and Petra Poklukar and Achin Kulshrestha and Amir Zamir and Federico Tombari},
  year = {2024},
  eprint = {2404.07204},
  archivePrefix = {arXiv},
  url = {https://arxiv.org/abs/2404.07204},
  note = {ECCV 2024}
}

@misc{wang2023biobridge,
  title = {{BioBridge: Bridging Biomedical Foundation Models via Knowledge Graphs}},
  author = {Zifeng Wang and Zichen Wang and Balasubramaniam Srinivasan and Vassilis N. Ioannidis and Huzefa Rangwala and Rishita Anubhai},
  year = {2023},
  eprint = {2310.03320},
  archivePrefix = {arXiv},
  url = {https://arxiv.org/abs/2310.03320},
  note = {ICLR 2024}
}

@misc{liu2023visual,
  title = {{Visual Instruction Tuning}},
  author = {Haotian Liu and Chunyuan Li and Qingyang Wu and Yong Jae Lee},
  year = {2023},
  eprint = {2304.08485},
  archivePrefix = {arXiv},
  url = {https://arxiv.org/abs/2304.08485},
  note = {NeurIPS 2023}
}

@misc{zhuang2024vector,
  title = {{Vector-ICL: In-context Learning with Continuous Vector Representations}},
  author = {Yufan Zhuang and Chandan Singh and Liyuan Liu and Jingbo Shang and Jianfeng Gao},
  year = {2024},
  eprint = {2410.05629},
  archivePrefix = {arXiv},
  url = {https://arxiv.org/abs/2410.05629}
}

@misc{driess2023palm,
  title = {{PaLM-E: An Embodied Multimodal Language Model}},
  author = {Danny Driess and Fei Xia and Mehdi S. M. Sajjadi and Corey Lynch and Aakanksha Chowdhery and Brian Ichter and Ayzaan Wahid and Jonathan Tompson and Quan Vuong and Tianhe Yu and Wenlong Huang and Yevgen Chebotar and Pierre Sermanet and Daniel Duckworth and Sergey Levine and Vincent Vanhoucke and Karol Hausman and Marc Toussaint and Klaus Greff and Andy Zeng and Igor Mordatch and Pete Florence},
  year = {2023},
  eprint = {2303.03378},
  archivePrefix = {arXiv},
  url = {https://arxiv.org/abs/2303.03378}
}

@misc{tsou2025bioverse,
  title = {{BioVERSE: Representation Alignment of Biomedical Modalities to LLMs for Multi-Modal Reasoning}},
  author = {Ching-Huei Tsou and Michal Ozery-Flato and Ella Barkan and Diwakar Mahajan and Ben Shapira},
  year = {2025},
  eprint = {2510.01428},
  archivePrefix = {arXiv},
  url = {https://arxiv.org/abs/2510.01428}
}

@misc{jin2023time,
  title = {{Time-LLM: Time Series Forecasting by Reprogramming Large Language Models}},
  author = {Ming Jin and Shiyu Wang and Lintao Ma and Zhixuan Chu and James Y. Zhang and Xiaoming Shi and Pin-Yu Chen and Yuxuan Liang and Yuan-Fang Li and Shirui Pan and Qingsong Wen},
  year = {2023},
  eprint = {2310.01728},
  archivePrefix = {arXiv},
  url = {https://arxiv.org/abs/2310.01728},
  note = {ICLR 2024}
}

@misc{tsimpoukelli2021multimodal,
  title = {{Multimodal Few-Shot Learning with Frozen Language Models}},
  author = {Maria Tsimpoukelli and Jacob Menick and Serkan Cabi and S. M. Ali Eslami and Oriol Vinyals and Felix Hill},
  year = {2021},
  eprint = {2106.13884},
  archivePrefix = {arXiv},
  url = {https://arxiv.org/abs/2106.13884},
  note = {NeurIPS 2021}
}

@misc{xiao2024proteingpt,
  title = {{ProteinGPT: Multimodal LLM for Protein Property Prediction and Structure Understanding}},
  author = {Yijia Xiao and Edward Sun and Yiqiao Jin and Qifan Wang and Wei Wang},
  year = {2024},
  eprint = {2408.11363},
  archivePrefix = {arXiv},
  url = {https://arxiv.org/abs/2408.11363},
  note = {MLGenX Workshop at ICLR 2025}
}

@misc{cheng2024xrag,
  title = {{xRAG: Extreme Context Compression for Retrieval-augmented Generation with One Token}},
  author = {Xin Cheng and Xun Wang and Xingxing Zhang and Tao Ge and Si-Qing Chen and Furu Wei and Huishuai Zhang and Dongyan Zhao},
  year = {2024},
  eprint = {2405.13792},
  archivePrefix = {arXiv},
  url = {https://arxiv.org/abs/2405.13792},
  note = {NeurIPS 2024}
}

@misc{liu2026beyond,
  title = {{Beyond tokens: a unified framework for latent communication in LLM-based multi-agent systems}},
  author = {Yingzhuo Liu},
  year = {2026},
  eprint = {2606.05711},
  archivePrefix = {arXiv},
  url = {https://arxiv.org/abs/2606.05711}
}

@misc{liu2026vision,
  title = {{The Vision Wormhole: Latent-Space Communication in Heterogeneous Multi-Agent Systems}},
  author = {Xiaoze Liu and Ruowang Zhang and Weichen Yu and Siheng Xiong and Liu He and Feijie Wu and Hoin Jung and Matt Fredrikson and Xiaoqian Wang and Jing Gao},
  year = {2026},
  eprint = {2602.15382},
  archivePrefix = {arXiv},
  url = {https://arxiv.org/abs/2602.15382}
}

@misc{chen2026see,
  title = {{See What I See, Know What I Think: Dense Latent Communication Across Heterogeneous Agents}},
  author = {Siyi Chen and Xiaoyan Zhang and Meng Wu and Jonathan Tremblay and Valts Blukis and Stan Birchfield and Rene Vidal and Alvaro Velasquez and Sijia Liu and Qing Qu},
  year = {2026},
  eprint = {2606.13594},
  archivePrefix = {arXiv},
  url = {https://arxiv.org/abs/2606.13594}
}

@misc{li2026heterogeneous,
  title = {{Heterogeneous Scientific Foundation Model Collaboration}},
  author = {Zihao Li and Jiaru Zou and Feihao Fang and Xuying Ning and Mengting Ai and Tianxin Wei and Sirui Chen and Xiyuan Yang and Jingrui He},
  year = {2026},
  eprint = {2604.27351},
  archivePrefix = {arXiv},
  url = {https://arxiv.org/abs/2604.27351}
}

@misc{rossi2026latent,
  title = {{Latent Cache Flow: Model-to-Model Communication Without Text}},
  author = {Maximillian Rossi and Prajwal Raghunath and Eugene Wu},
  year = {2026},
  eprint = {2605.22863},
  archivePrefix = {arXiv},
  url = {https://arxiv.org/abs/2605.22863}
}

@article{almeida2025multimodal,
  title = {{A multimodal conversational agent for DNA, RNA and protein tasks}},
  author = {Bernardo P. de Almeida and Guillaume Richard and Hugo Dalla-Torre and Christopher Blum and Lorenz Hexemer and Priyanka Pandey and Stefan Laurent and Chandana Rajesh and Marie Lopez and Alexandre Laterre and Maren Lang and Uğur Şahin and Karim Beguir and Thomas Pierrot},
  year = {2025},
  journal = {Nature Machine Intelligence},
  volume = {7},
  pages = {928--941},
  doi = {10.1038/s42256-025-01047-1}
}

@article{boiko2023autonomous,
  title = {{Autonomous chemical research with large language models}},
  author = {Daniil A. Boiko and Robert MacKnight and Ben Kline and Gabe Gomes},
  year = {2023},
  journal = {Nature},
  volume = {624},
  pages = {570--578},
  doi = {10.1038/s41586-023-06792-0}
}

@article{bran2024augmenting,
  title = {{Augmenting large language models with chemistry tools}},
  author = {Andres M. Bran and Sam Cox and Oliver Schilter and Carlo Baldassari and Andrew D. White and Philippe Schwaller},
  year = {2024},
  journal = {Nature Machine Intelligence},
  volume = {6},
  pages = {525--535},
  doi = {10.1038/s42256-024-00832-8}
}

@misc{zhou2025decoding,
  title = {{Decoding the Molecular Language of Proteins with Evolla}},
  author = {Xibin Zhou and Chenchen Han and Yingqi Zhang and Huan Du and Jiayuan Tian and Jin Su and Renju Liu and Kai Zhuang and Shiyu Jiang and Anthony Gitter and Li Liu and Huayu Li and Meiqi Wu and Shiyang You and Zichen Yuan and Feng Ju and Huilin Zhang and Wei Zheng and Fengyuan Dai and Yuyang Zhou and Yuyang Tao and Dan Wu and Zongze Shao and Yang Liu and Hongyuan Lu and Fajie Yuan},
  year = {2025},
  howpublished = {bioRxiv preprint},
  doi = {10.1101/2025.01.05.630192}
}

@misc{huang2025biomni,
  title = {{Biomni: A General-Purpose Biomedical AI Agent}},
  author = {Kexin Huang and Serena Zhang and Hanchen Wang and Yuanhao Qu and Yingzhou Lu and Yusuf Roohani and Ryan Li and Lin Qiu and Gavin Li and Junze Zhang and Di Yin and Shruti Marwaha and Jennefer N. Carter and Xin Zhou and Matthew Wheeler and Jonathan A. Bernstein and Mengdi Wang and Peng He and Jingtian Zhou and Michael Snyder and Le Cong and Aviv Regev and Jure Leskovec},
  year = {2025},
  howpublished = {bioRxiv preprint},
  doi = {10.1101/2025.05.30.656746}
}

@misc{abdine2023prot2text,
  title = {{Prot2Text: Multimodal Protein's Function Generation with GNNs and Transformers}},
  author = {Hadi Abdine and Michail Chatzianastasis and Costas Bouyioukos and Michalis Vazirgiannis},
  year = {2023},
  eprint = {2307.14367},
  archivePrefix = {arXiv},
  url = {https://arxiv.org/abs/2307.14367},
  note = {AAAI 2024, 38(10):10757--10765, doi:10.1609/aaai.v38i10.28948}
}

@misc{wang2023instructprotein,
  title = {{InstructProtein: Aligning Human and Protein Language via Knowledge Instruction}},
  author = {Zeyuan Wang and Qiang Zhang and Keyan Ding and Ming Qin and Xiang Zhuang and Xiaotong Li and Huajun Chen},
  year = {2023},
  eprint = {2310.03269},
  archivePrefix = {arXiv},
  url = {https://arxiv.org/abs/2310.03269}
}

@misc{zhuo2024protllm,
  title = {{ProtLLM: An Interleaved Protein-Language LLM with Protein-as-Word Pre-Training}},
  author = {Le Zhuo and Zewen Chi and Minghao Xu and Heyan Huang and Heqi Zheng and Conghui He and Xian-Ling Mao and Wentao Zhang},
  year = {2024},
  eprint = {2403.07920},
  archivePrefix = {arXiv},
  url = {https://arxiv.org/abs/2403.07920}
}

@misc{liu2023molca,
  title = {{MolCA: Molecular Graph-Language Modeling with Cross-Modal Projector and Uni-Modal Adapter}},
  author = {Zhiyuan Liu and Sihang Li and Yanchen Luo and Hao Fei and Yixin Cao and Kenji Kawaguchi and Xiang Wang and Tat-Seng Chua},
  year = {2023},
  eprint = {2310.12798},
  archivePrefix = {arXiv},
  url = {https://arxiv.org/abs/2310.12798},
  note = {EMNLP 2023}
}

@misc{li2024towards,
  title = {{Towards 3D Molecule-Text Interpretation in Language Models}},
  author = {Sihang Li and Zhiyuan Liu and Yanchen Luo and Xiang Wang and Xiangnan He and Kenji Kawaguchi and Tat-Seng Chua and Qi Tian},
  year = {2024},
  eprint = {2401.13923},
  archivePrefix = {arXiv},
  url = {https://arxiv.org/abs/2401.13923}
}

@misc{cao2023instructmol,
  title = {{InstructMol: Multi-Modal Integration for Building a Versatile and Reliable Molecular Assistant in Drug Discovery}},
  author = {He Cao and Zijing Liu and Xingyu Lu and Yuan Yao and Yu Li},
  year = {2023},
  eprint = {2311.16208},
  archivePrefix = {arXiv},
  url = {https://arxiv.org/abs/2311.16208}
}

@misc{pei2023biot5,
  title = {{BioT5: Enriching Cross-modal Integration in Biology with Chemical Knowledge and Natural Language Associations}},
  author = {Qizhi Pei and Wei Zhang and Jinhua Zhu and Kehan Wu and Kaiyuan Gao and Lijun Wu and Yingce Xia and Rui Yan},
  year = {2023},
  eprint = {2310.07276},
  archivePrefix = {arXiv},
  url = {https://arxiv.org/abs/2310.07276},
  note = {EMNLP 2023}
}

@misc{fallahpour2025bioreason,
  title = {{BioReason: Incentivizing Multimodal Biological Reasoning within a DNA-LLM Model}},
  author = {Adibvafa Fallahpour and Andrew Magnuson and Purav Gupta and Shihao Ma and Jack Naimer and Arnav Shah and Haonan Duan and Omar Ibrahim and Hani Goodarzi and Chris J. Maddison and Bo Wang},
  year = {2025},
  eprint = {2505.23579},
  archivePrefix = {arXiv},
  url = {https://arxiv.org/abs/2505.23579}
}

@misc{xia2025nature,
  title = {{Nature Language Model: Deciphering the Language of Nature for Scientific Discovery}},
  author = {Yingce Xia and Peiran Jin and Shufang Xie and Liang He and Chuan Cao and Renqian Luo and Guoqing Liu and Yue Wang and Zequn Liu and Yuan-Jyue Chen and others},
  year = {2025},
  eprint = {2502.07527},
  archivePrefix = {arXiv},
  url = {https://arxiv.org/abs/2502.07527}
}

@misc{bai2025intern,
  title = {{Intern-S1: A Scientific Multimodal Foundation Model}},
  author = {Lei Bai and Zhongrui Cai and Yuhang Cao and Maosong Cao and Weihan Cao and Chiyu Chen and Haojiong Chen and Kai Chen and Pengcheng Chen and Ying Chen and others},
  year = {2025},
  eprint = {2508.15763},
  archivePrefix = {arXiv},
  url = {https://arxiv.org/abs/2508.15763}
}

@misc{wang2025txgemma,
  title = {{TxGemma: Efficient and Agentic LLMs for Therapeutics}},
  author = {Eric Wang and Samuel Schmidgall and Paul F. Jaeger and Fan Zhang and Rory Pilgrim and Yossi Matias and Joelle Barral and David Fleet and Shekoofeh Azizi},
  year = {2025},
  eprint = {2504.06196},
  archivePrefix = {arXiv},
  url = {https://arxiv.org/abs/2504.06196}
}

@misc{luo2023biomedgpt,
  title = {{BioMedGPT: Open Multimodal Generative Pre-trained Transformer for BioMedicine}},
  author = {Yizhen Luo and Jiahuan Zhang and Siqi Fan and Kai Yang and Yushuai Wu and Mu Qiao and Zaiqing Nie},
  year = {2023},
  eprint = {2308.09442},
  archivePrefix = {arXiv},
  url = {https://arxiv.org/abs/2308.09442}
}

@article{schaefer2025multimodal,
  title = {{Multimodal learning enables chat-based exploration of single-cell data}},
  author = {Moritz Schaefer and Peter Peneder and Daniel Malzl and Salvo Danilo Lombardo and Mihaela Peycheva and Jake Burton and Anna Hakobyan and Varun Sharma and Thomas Krausgruber and Celine Sin and J{\"o}rg Menche and Eleni M. Tomazou and Christoph Bock},
  year = {2025},
  journal = {Nature Biotechnology},
  doi = {10.1038/s41587-025-02857-9},
  note = {CellWhisperer; preprint doi:10.1101/2024.10.15.618501}
}

@inproceedings{levine2024cell2sentence,
  title = {{Cell2Sentence: Teaching Large Language Models the Language of Biology}},
  author = {Daniel Levine and Syed Asad Rizvi and Sacha L{\'e}vy and Nazreen Pallikkavaliyaveetil and David Zhang and Xingyu Chen and Sina Ghadermarzi and Ruiming Wu and Zihe Zheng and Ivan Vrkic and Anna Zhong and Daphne Raskin and Insu Han and Antonio Henrique De Oliveira Fonseca and Josue Ortega Caro and Amin Karbasi and Rahul Madhav Dhodapkar and David van Dijk},
  booktitle = {Proceedings of the 41st International Conference on Machine Learning},
  series = {PMLR},
  volume = {235},
  pages = {27299--27325},
  year = {2024},
  note = {Preprint doi:10.1101/2023.09.11.557287}
}

@misc{rizvi2025scaling,
  title = {{Scaling Large Language Models for Next-Generation Single-Cell Analysis}},
  author = {Syed Asad Rizvi and Daniel Levine and Aakash Patel and Shiyang Zhang and Eric Wang and Curtis Jamison Perry and Ivan Vrkic and Nicole Mayerli Constante and Zirui Fu and Sizhuang He and David Zhang and Cerise Tang and Zhuoyang Lyu and Rayyan Darji and Chang Li and Emily Sun and David Jeong and Lawrence Zhao and Jennifer Kwan and David Braun and Brian Hafler and Hattie Chung and Rahul M. Dhodapkar and Paul Jaeger and Bryan Perozzi and Jeffrey Ishizuka and Shekoofeh Azizi and David van Dijk},
  year = {2025},
  howpublished = {bioRxiv preprint},
  doi = {10.1101/2025.04.14.648850},
  note = {C2S-Scale}
}

@misc{han2023imagebindllm,
  title = {{ImageBind-LLM: Multi-modality Instruction Tuning}},
  author = {Jiaming Han and Renrui Zhang and Wenqi Shao and Peng Gao and Peng Xu and Han Xiao and Kaipeng Zhang and Chris Liu and Song Wen and Ziyu Guo and Xudong Lu and Shuai Ren and Yafei Wen and Xiaoxin Chen and Xiangyu Yue and Hongsheng Li and Yu Qiao},
  year = {2023},
  eprint = {2309.03905},
  archivePrefix = {arXiv},
  url = {https://arxiv.org/abs/2309.03905}
}

@misc{han2023onellm,
  title = {{OneLLM: One Framework to Align All Modalities with Language}},
  author = {Jiaming Han and Kaixiong Gong and Yiyuan Zhang and Jiaqi Wang and Kaipeng Zhang and Dahua Lin and Yu Qiao and Peng Gao and Xiangyu Yue},
  year = {2023},
  eprint = {2312.03700},
  archivePrefix = {arXiv},
  url = {https://arxiv.org/abs/2312.03700},
  note = {CVPR 2024}
}

@misc{hao2022metalm,
  title = {{Language Models are General-Purpose Interfaces}},
  author = {Yaru Hao and Haoyu Song and Li Dong and Shaohan Huang and Zewen Chi and Wenhui Wang and Shuming Ma and Furu Wei},
  year = {2022},
  eprint = {2206.06336},
  archivePrefix = {arXiv},
  url = {https://arxiv.org/abs/2206.06336},
  note = {MetaLM}
}

@misc{zeng2022socratic,
  title = {{Socratic Models: Composing Zero-Shot Multimodal Reasoning with Language}},
  author = {Andy Zeng and Maria Attarian and Brian Ichter and Krzysztof Choromanski and Adrian Wong and Stefan Welker and Federico Tombari and Aveek Purohit and Michael Ryoo and Vikas Sindhwani and Johnny Lee and Vincent Vanhoucke and Pete Florence},
  year = {2022},
  eprint = {2204.00598},
  archivePrefix = {arXiv},
  url = {https://arxiv.org/abs/2204.00598},
  note = {ICLR 2023}
}

@misc{shen2023hugginggpt,
  title = {{HuggingGPT: Solving AI Tasks with ChatGPT and its Friends in Hugging Face}},
  author = {Yongliang Shen and Kaitao Song and Xu Tan and Dongsheng Li and Weiming Lu and Yueting Zhuang},
  year = {2023},
  eprint = {2303.17580},
  archivePrefix = {arXiv},
  url = {https://arxiv.org/abs/2303.17580},
  note = {NeurIPS 2023}
}

@misc{bansal2024calm,
  title = {{LLM Augmented LLMs: Expanding Capabilities through Composition}},
  author = {Rachit Bansal and Bidisha Samanta and Siddharth Dalmia and Nitish Gupta and Shikhar Vashishth and Sriram Ganapathy and Abhishek Bapna and Prateek Jain and Partha Talukdar},
  year = {2024},
  eprint = {2401.02412},
  archivePrefix = {arXiv},
  url = {https://arxiv.org/abs/2401.02412},
  note = {ICLR 2024; CALM}
}

@misc{ghandeharioun2024patchscopes,
  title = {{Patchscopes: A Unifying Framework for Inspecting Hidden Representations of Language Models}},
  author = {Asma Ghandeharioun and Avi Caciularu and Adam Pearce and Lucas Dixon and Mor Geva},
  year = {2024},
  eprint = {2401.06102},
  archivePrefix = {arXiv},
  url = {https://arxiv.org/abs/2401.06102},
  note = {ICML 2024}
}

@misc{moschella2023relative,
  title = {{Relative representations enable zero-shot latent space communication}},
  author = {Luca Moschella and Valentino Maiorca and Marco Fumero and Antonio Norelli and Francesco Locatello and Emanuele Rodol{\`a}},
  year = {2022},
  eprint = {2209.15430},
  archivePrefix = {arXiv},
  url = {https://arxiv.org/abs/2209.15430},
  note = {ICLR 2023}
}

@misc{maiorca2023latent,
  title = {{Latent Space Translation via Semantic Alignment}},
  author = {Valentino Maiorca and Luca Moschella and Antonio Norelli and Marco Fumero and Francesco Locatello and Emanuele Rodol{\`a}},
  year = {2023},
  eprint = {2311.00664},
  archivePrefix = {arXiv},
  url = {https://arxiv.org/abs/2311.00664},
  note = {NeurIPS 2023}
}

@misc{turner2023actadd,
  title = {{Steering Language Models With Activation Engineering}},
  author = {Alexander Matt Turner and Lisa Thiergart and Gavin Leech and David Udell and Juan J. Vazquez and Ulisse Mini and Monte MacDiarmid},
  year = {2023},
  eprint = {2308.10248},
  archivePrefix = {arXiv},
  url = {https://arxiv.org/abs/2308.10248},
  note = {ActAdd; earlier title: Activation Addition: Steering Language Models Without Optimization}
}

@misc{zhang2024wings,
  title = {{Wings: Learning Multimodal LLMs without Text-only Forgetting}},
  author = {Yi-Kai Zhang and Shiyin Lu and Yang Li and Yanqing Ma and Qing-Guo Chen and Zhao Xu and Weihua Luo and Kaifu Zhang and De-Chuan Zhan and Han-Jia Ye},
  year = {2024},
  eprint = {2406.03496},
  archivePrefix = {arXiv},
  url = {https://arxiv.org/abs/2406.03496},
  note = {NeurIPS 2024}
}

@misc{anthropic2024mcp,
  title        = {Introducing the {Model Context Protocol}},
  author       = {{Anthropic}},
  year         = {2024},
  month        = nov,
  howpublished = {\url{https://www.anthropic.com/news/model-context-protocol}},
  note         = {Announcement, 25 November 2024. Specification: \url{https://modelcontextprotocol.io/specification/2025-06-18}}
}

@article{uniprot2025uniprot,
  title = {{UniProt: the Universal Protein Knowledgebase in 2025}},
  author = {{The UniProt Consortium}},
  year = {2025},
  journal = {Nucleic Acids Research},
  volume = {53},
  number = {D1},
  pages = {D609--D617},
  doi = {10.1093/nar/gkae1010}
}

@article{bansal2022rhea,
  title = {{Rhea, the reaction knowledgebase in 2022}},
  author = {Parit Bansal and Anne Morgat and Kristian B. Axelsen and Venkatesh Muthukrishnan and Elisabeth Coudert and Lucila Aimo and Nevila Hyka-Nouspikel and Elisabeth Gasteiger and Arnaud Kerhornou and Teresa Batista Neto and Monica Pozzato and Marie-Claude Blatter and Alex Ignatchenko and Nicole Redaschi and Alan Bridge},
  year = {2022},
  journal = {Nucleic Acids Research},
  volume = {50},
  number = {D1},
  pages = {D693--D700},
  doi = {10.1093/nar/gkab1016}
}

@article{malik2026chebi,
  title = {{ChEBI: re-engineered for a sustainable future}},
  author = {Adnan Malik and Muhammad Arsalan and Carlos Moreno and Juan Mosquera and Eloy F{\'e}lix and Tevfik Kizil{\"o}ren and Venkatesh Muthukrishnan and Barbara Zdrazil and Andrew R. Leach and Noel M. O'Boyle},
  year = {2026},
  journal = {Nucleic Acids Research},
  volume = {54},
  number = {D1},
  pages = {D1768--D1778},
  doi = {10.1093/nar/gkaf1271}
}

@article{steinegger2017mmseqs2,
  title = {{MMseqs2 enables sensitive protein sequence searching for the analysis of massive data sets}},
  author = {Martin Steinegger and Johannes S{\"o}ding},
  year = {2017},
  journal = {Nature Biotechnology},
  volume = {35},
  number = {11},
  pages = {1026--1028},
  doi = {10.1038/nbt.3988}
}

@misc{golkar2026mimic,
  title = {{MIMIC: A Generative Multimodal Foundation Model for Biomolecules}},
  author = {Siavash Golkar and Jake Kovalic and Irina Espejo Morales and Samuel Sledzieski and Minhuan Li and Ksenia Sokolova and Geraud Krawezik and Alberto Bietti and Claudia Skok Gibbs and Roman Klypa and Shengwei Xiong and Francois Lanusse and Liam Parker and Kyunghyun Cho and Miles Cranmer and Tom Hehir and Michael McCabe and Lucas Meyer and Rudy Morel and Payel Mukhopadhyay and Mariel Pettee and Helen Qu and Jeff Shen and David Fouhey and Hadi Sotoudeh and Vikram Mulligan and Pilar Cossio and Sonya M. Hanson and Alisha N. Jones and Olga G. Troyanskaya and Shirley Ho},
  year = {2026},
  eprint = {2604.24506},
  archivePrefix = {arXiv},
  url = {https://arxiv.org/abs/2604.24506}
}

@misc{candido2026language,
  title = {{Language Modeling Materializes a World Model of Protein Biology}},
  author = {Salvatore Candido and Thomas Hayes and Alexander Derry and Roshan Rao and Zeming Lin and others},
  year = {2026},
  howpublished = {bioRxiv preprint},
  doi = {10.64898/2026.06.03.729735},
  url = {https://www.biorxiv.org/content/10.64898/2026.06.03.729735},
  note = {ESM C}
}

@article{chen2025xtrimopglm,
  title = {{xTrimoPGLM: unified 100-billion-parameter pretrained transformer for deciphering the language of proteins}},
  author = {Bo Chen and Xingyi Cheng and Pan Li and Yangli-ao Geng and Jing Gong and Shen Li and Zhilei Bei and Xu Tan and Boyan Wang and Xin Zeng and Chiming Liu and Aohan Zeng and Yuxiao Dong and Jie Tang and Le Song},
  year = {2025},
  journal = {Nature Methods},
  volume = {22},
  number = {5},
  pages = {1028--1039},
  doi = {10.1038/s41592-025-02636-z}
}

@article{ross2022large,
  title = {{Large-scale chemical language representations capture molecular structure and properties}},
  author = {Jerret Ross and Brian Belgodere and Vijil Chenthamarakshan and Inkit Padhi and Youssef Mroueh and Payel Das},
  year = {2022},
  journal = {Nature Machine Intelligence},
  volume = {4},
  number = {12},
  pages = {1256--1264},
  doi = {10.1038/s42256-022-00580-7},
  note = {MoLFormer}
}

@misc{ji2024unimol2,
  title = {{Uni-Mol2: Exploring Molecular Pretraining Model at Scale}},
  author = {Xiaohong Ji and Zhen Wang and Zhifeng Gao and Hang Zheng and Linfeng Zhang and Guolin Ke and Weinan E},
  year = {2024},
  eprint = {2406.14969},
  archivePrefix = {arXiv},
  url = {https://arxiv.org/abs/2406.14969}
}

@article{cai2025chemfm,
  title = {{ChemFM as a scaling law guided foundation model pre-trained on informative chemicals}},
  author = {Feiyang Cai and Katelin Zacour and Tianyu Zhu and Tzuen-Rong Tzeng and Yongping Duan and Ling Liu and Srikanth Pilla and Gang Li and Feng Luo},
  year = {2025},
  journal = {Communications Chemistry},
  volume = {9},
  pages = {3},
  doi = {10.1038/s42004-025-01793-8},
  note = {Preprint arXiv:2410.21422}
}

@misc{kimiteam2025kimik2,
  author = {{Kimi Team}},
  title = {{Kimi K2: Open Agentic Intelligence}},
  year = {2025},
  eprint = {2507.20534},
  archivePrefix = {arXiv},
  url = {https://arxiv.org/abs/2507.20534}
}

@misc{moonshot2026kimik26,
  author = {{Moonshot AI}},
  title = {{Kimi K2.6} Model Card},
  year = {2026},
  howpublished = {\url{https://huggingface.co/moonshotai/Kimi-K2.6}},
  note = {Model Summary; accessed September 29, 2026}
}

@misc{golkar2024xval,
  title = {{xVal}: A Continuous Numerical Tokenization for Scientific Language Models},
  author = {Siavash Golkar and Mariel Pettee and Michael Eickenberg and Alberto Bietti and Miles Cranmer and Geraud Krawezik and Fran\c{c}ois Lanusse and Michael McCabe and Ruben Ohana and Liam Parker and Bruno R\'{e}galdo-Saint Blancard and Tiberiu Tesileanu and Kyunghyun Cho and Shirley Ho},
  year = {2024},
  eprint = {2310.02989},
  archivePrefix = {arXiv},
  primaryClass = {stat.ML},
  url = {https://arxiv.org/abs/2310.02989}
}

@misc{hao2024coconut,
  title = {Training Large Language Models to Reason in a Continuous Latent Space},
  author = {Shibo Hao and Sainbayar Sukhbaatar and DiJia Su and Xian Li and Zhiting Hu and Jason Weston and Yuandong Tian},
  year = {2024},
  eprint = {2412.06769},
  archivePrefix = {arXiv},
  primaryClass = {cs.CL},
  url = {https://arxiv.org/abs/2412.06769}
}

@misc{li2025latentvisual,
  title = {Latent Visual Reasoning},
  author = {Bangzheng Li and Ximeng Sun and Jiang Liu and Ze Wang and Jialian Wu and Xiaodong Yu and Hao Chen and Emad Barsoum and Muhao Chen and Zicheng Liu},
  year = {2025},
  eprint = {2509.24251},
  archivePrefix = {arXiv},
  primaryClass = {cs.CV},
  url = {https://arxiv.org/abs/2509.24251}
}

@misc{foote2026trellis,
  author = {Foote, Addie and Laine, Rudolf and Kostolansky, Tim},
  title = {Post-Training 50x Faster},
  year = {2026},
  month = mar,
  howpublished = {Workshop Labs, \url{https://www.workshoplabs.ai/blog/post-training-50x-faster}},
  note = {March 12, 2026}
}
